\documentclass{article}
\usepackage{iclr2027_conference}
\iclrfinalcopy

\usepackage{hyperref}
\usepackage{url}
\usepackage{graphicx}
\usepackage{caption}
\usepackage{booktabs}
\usepackage{algorithm}
\usepackage{algpseudocode}
\usepackage{amsmath,amssymb,mathtools}

\usepackage{tikz,xcolor,array}
\usepackage{pgf}
\usetikzlibrary{arrows.meta,calc,positioning,fit,backgrounds,spy, matrix}
\title{Mesh2Shell Neural Operator for Inverse Shape Optimization}

\author{Xinyi Li\textsuperscript{*}, Xi Deng\textsuperscript{*}, Yubo Cao, Hudson Liu, Anima Anandkumar\\[0.5em]
{\small Computing and Mathematical Sciences, California Institute of Technology}\\[0.25em]
{\small \textsuperscript{*}Equal contribution}}\date{}

\usepackage{booktabs}
\usepackage{pifont}
\usepackage{array}
\usepackage{graphicx}

\begin{document}

\maketitle

\begin{abstract}
Physics-driven design seeks high-performing geometries, which need not
conform to a learned shape category. Parametric and latent-space approaches
facilitate surrogate-based optimization, but restrict deformations to
prescribed shape families or learned decoders.
We introduce the \textbf{M}esh-to-\textbf{S}hell \textbf{N}eural
\textbf{O}perator (M2S-NO) as the mesh-to-shell component of a
category-agnostic inverse shape optimization framework.
We separate the explicit mesh used for design from the volumetric
representation used for physics prediction. M2S-NO
maps oriented surface samples to a surface-centered \emph{shell
representation}, such as TruncatedSDF or HeatSDF. A separate problem-specific
PDE neural operator (PDE-NO) maps this representation to physical fields.
Objective gradients update mesh vertices
through both operators, without a latent shape decoder. Preconditioning and
remeshing support free-form deformation, while fixed interfaces are controlled
directly on the mesh. Optional shell-space proposals enable topology changes.
We evaluate geometric matching, jet-engine brackets, and neutron moderators,
with independent FEM and OpenMC evaluation. Relative to the best training
designs, M2S-NO reduces
maximum bracket displacement by $14.3\%$ and increases neutron-sensor current
by $13.8\%$. Optimization also succeeds from an out-of-distribution bracket
initialization, demonstrating the framework's geometric flexibility.

\end{abstract}

\section{Introduction}
In physics-driven design, the desired behavior can be specified before the
shape is known. A bracket must connect prescribed interfaces and resist
deformation; a neutron moderator must surround a sensor and improve its
response. Neither requirement determines what the shape should look like.
We study \emph{category-agnostic inverse shape optimization}: physical
objectives and geometric constraints guide the search without a prescribed
shape family or a learned category-specific decoder. Neural operators
accelerate physical evaluation~\citep{li2021fno,li2023geofno,li2023gino}, but
the design representation determines which geometric changes can be explored.

Existing approaches restrict the design space in two ways. Parametric methods
optimize prescribed deformation variables; Geodesic Convolutional Shape Optimization
uses 21 parameters for car design~\citep{baque2018geodesic}. Learned shape
spaces offer richer deformations, but updates still act through a decoder.
MeshSDF differentiates through an extracted mesh to update a
latent code~\citep{remelli2020meshsdf}, GANO optimizes an SDF
auto-decoder's code~\citep{sun2026gano}, and generative methods combine
physical guidance with learned shape distributions~\citep{you2025physgen,chen2026aerodynamic}.
These representations regularize optimization but can exclude useful
deformations or poorly represent a new starting geometry. We seek direct
geometric control without such a restricted parameterization.

Unlike these approaches, an explicit mesh keeps the design in geometric space
rather than a prescribed deformation basis or category-specific learned shape
space. It exposes individual vertices for
optimization, preserves the input geometry without decoder reconstruction, and
allows interfaces to remain fixed. However, unconstrained vertex-wise updates
can exploit learned physics predictors: MeshSDF's vertex-wise baseline achieved
low predicted drag but produced implausible geometries with nonconvergent CFD
simulations~\citep{remelli2020meshsdf}. Volumetric fields suit grid-based
physics surrogates, but direct field optimization makes surface control
indirect and leaves a large, weakly constrained design space. The challenge is
to combine mesh-space freedom with an effective gradient pathway through the
physics predictor. We do so by separating the representations used to
\emph{modify geometry} and \emph{predict its physics}.

\begin{figure*}[t]
    \centering
    \input{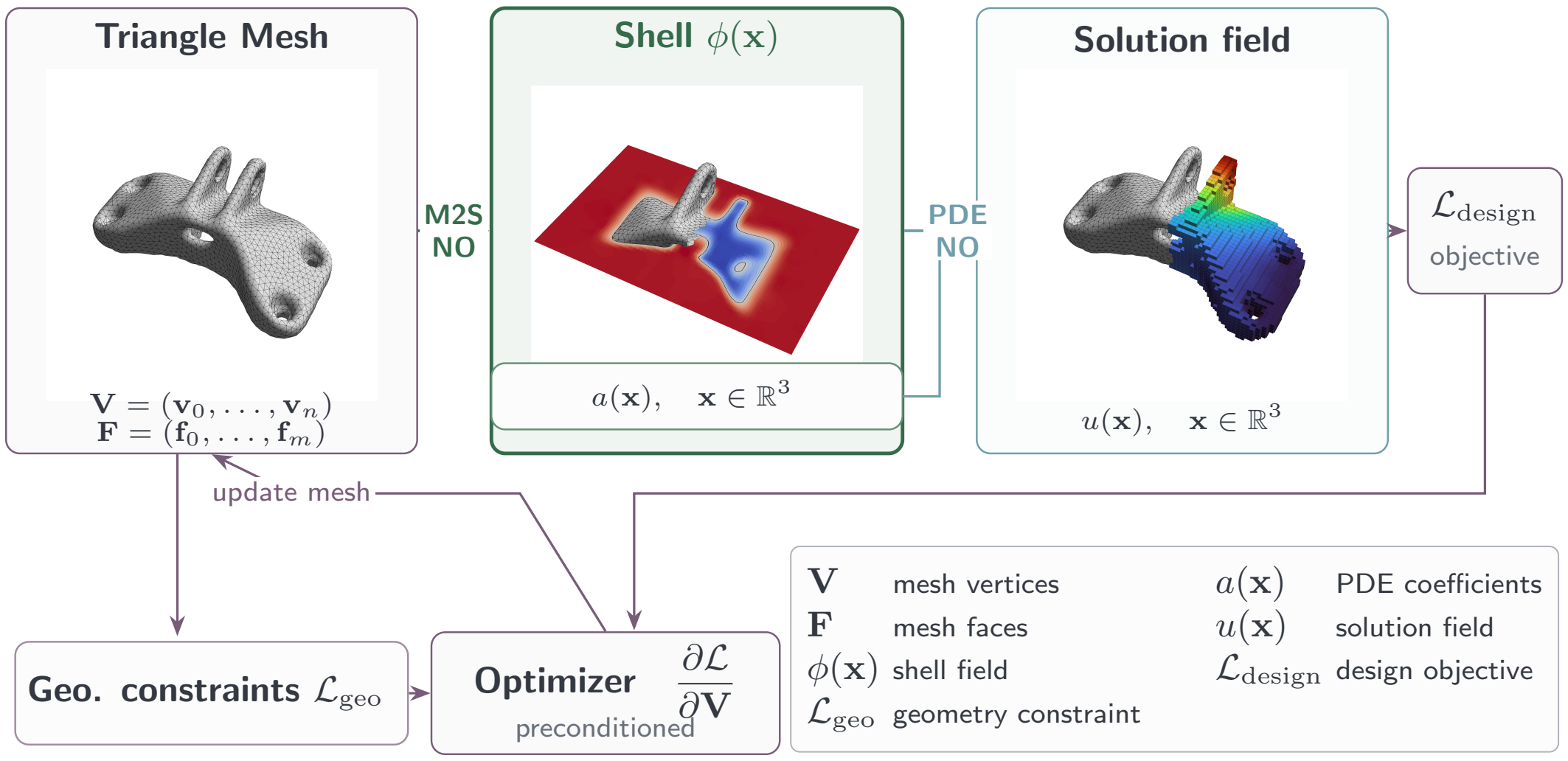}
    \caption{\textbf{M2S-NO-based inverse shape optimization.} The mesh is the design
    variable and the shell interfaces with volumetric solution-field prediction. Gradients propagate
    through both operators to mesh vertices, followed by geometric control and
    preconditioned updates.}
    \label{fig:pipeline overview}
\end{figure*}

We introduce the \textbf{Mesh-to-Shell Neural Operator (M2S-NO}), a learned
mesh-to-shell operator providing a surface-to-volume interface for direct
mesh-based design (Figure~\ref{fig:pipeline overview}). M2S-NO maps oriented
mesh samples to a surface-centered \emph{shell representation}, instantiated
by TruncatedSDF or HeatSDF. A separate PDE neural operator (PDE-NO) maps the
shell and physical inputs to a solution field. Crucially, M2S-NO
\emph{encodes the supplied surface} rather than generating it from a latent
shape code. Together, these operators form an inverse shape optimization
framework in which the mesh remains the design variable and objective
gradients propagate through both. M2S-NO is pretrained on diverse procedural
shapes, separately from each problem's PDE-NO.

During inverse optimization, gradients of the physical objective are
backpropagated through the PDE-NO and M2S-NO to the mesh vertices. A
mesh-Laplacian preconditioner smooths these updates, suppresses unstable, high-frequency deformations. When an edge-length check
detects a degraded mesh, remeshing restores the discretization without
retraining M2S-NO, whose input is a set of oriented samples rather than a fixed-size mesh. For topology changes, an optional shell-space step updates the predicted
shell, extracts a candidate surface and
accepts it only if it changes topology and lowers the full mesh-domain loss.

We evaluate free-form deformation and topology change via geometric matching,
and physical design through DeepJEB
brackets~\citep{hong2025deepjeb} and neutron
moderators~\citep{deng2026differentiable}. Shell-based variants outperform
the direct surface-to-field ablation in both physics tasks. M2S-NO improves
over the training-set best by $14.3\%$ in maximum bracket displacement and
$13.8\%$ in neutron-sensor current. An out-of-distribution, arch-augmented
bracket initialization also reaches a displacement $13.4\%$ below the
training-set best. Independent FEM and OpenMC evaluations validate the
optimized outcomes and the benefit of the shell interface.

Our contributions are:

\noindent\makebox[1.1em][l]{\textbullet}\textbf{Category-agnostic mesh design.} Direct surface optimization
through learned physics, with local control and without a prescribed shape
family or learned decoder.
\par\noindent\makebox[1.1em][l]{\textbullet}\textbf{A learned mesh-to-shell interface.} A geometry-supervised
interface connecting evolving meshes to volumetric physics, compatible with
remeshing and shell-space topology proposals.
\par\noindent\makebox[1.1em][l]{\textbullet}\textbf{Solver-evaluated optimization.} Improvements beyond the
best training designs in elasticity and neutron transport, including an OOD
bracket initialization.

\section{Related Work}
\paragraph{Geometry-aware neural operators.}
FNO uses regular-grid spectral convolutions~\citep{li2021fno}; geometry-aware
extensions use deformations, graph lifting, optimal transport, or localized
kernels~\citep{li2023geofno,li2023gino,li2025otno,li2024localno}.
M2S-NO supervises an intermediate shell field, separating mesh-to-shell
conversion from physics prediction.

\paragraph{Design variables and geometric control.}
Using a mesh as surrogate input does not imply optimizing its vertices
independently. \citet{baque2018geodesic} optimize 21 deformation parameters for
cars; \citet{rehmann2025surrogate} optimize rounded-cone parameters through
an SDF-based predictor. Mollified GNO enables geometric derivatives and
demonstrates airfoil optimization through spline control
nodes~\citep{lin2025mgno}. Learned simulators also support OOD design of
articulated tools and high-dimensional height fields~\citep{allen2022physical}.
Concurrent work CANTO~\citep{leibovici2026canto} combines NURBS-native geometry
encoding with an AB-UPT-based solution branch, demonstrating inverse design
within a six-parameter Ahmed-body CAD family.
Our free-form closed surface meshes with evolving discretizations permit
shape updates beyond prescribed CAD parameterizations.
Level-set optimization instead updates implicit
fields~\citep{oellerich2025levelset}. We retain explicit surface control and
use the preconditioner of \citet{nicolet2021large} to regularize mesh motion.

\paragraph{Learned shape spaces and generative design.}
DeepSDF represents shapes with latent codes and an implicit
decoder~\citep{park2019deepsdf}. MeshSDF differentiates iso-surface extraction
to update those codes, rather than independently optimizing the extracted
vertices~\citep{remelli2020meshsdf}. Its separate vertex-wise baseline exposes
the difficulty of unregularized mesh optimization. GANO likewise updates
latent codes and uses null-space projection for first-order preservation of
local constraints~\citep{sun2026gano}. DiffAirfoil, PhysGen, and related
aerodynamic methods incorporate learned shape distributions through diffusion,
flow matching, or distributional guidance
\citep{wei2024diffairfoil,you2025physgen,chen2026aerodynamic}.
These priors favor plausible shapes. M2S-NO instead encodes an explicit
design for physical evaluation without restricting its updates to a decoder's
range; regularization and fixed-interface control act in mesh space.

\section{Preliminaries}
\label{sec:preliminaries}

\paragraph{Distance fields and shells.}
A shell is a surface-centered volumetric representation whose geometric
variation is concentrated near the interface, rather than a thin physical
solid. The signed distance function (SDF) of $\mathcal S$ is
\begin{equation}
 \phi_{\mathrm{SDF}}(x)=s_{\mathcal S}(x)\min_{p\in\mathcal S}\|x-p\|_2,
 \qquad x\in\Omega,
 \label{eq:prelim-sdf}
\end{equation}
with $s_{\mathcal S}=-1$ inside and $+1$ outside.  A truncated SDF,
\begin{equation}
 \phi_{\mathrm{TSDF}}=\operatorname{clip}(\phi_{\mathrm{SDF}},-\delta,\delta),
 \label{eq:prelim-tsdf}
\end{equation}
is constant outside a band of width $\delta$ on either side of the surface.
Thus, TruncatedSDF is a shell when $\delta$ is small, whereas an untruncated
SDF is not.

HeatSDF is a diffusion-smoothed field and need not be an exact signed distance
function. To construct its offline targets, oriented surface samples $(P,N)$
are splatted to a grid normal field $X^{(0)}$, diffused, normalized, and used
to reconstruct a scalar potential:
\begin{equation}
 \begin{aligned}
 X^{(q+1)}&=(I-\tau\Delta_h)^{-1}X^{(q)},
 &&q=0,\ldots,K-1,\\[-2pt]
 Y_i&=\frac{X_i^{(K)}}{\|X_i^{(K)}\|_2+\epsilon_Y},
 &\Delta_h\phi&=\nabla_h\!\cdot Y.
 \end{aligned}
 \label{eq:prelim-heatsdf}
\end{equation}
Here $\Delta_h$ is the grid Laplacian, $\tau$ is the per-step diffusion size,
and $K$ is the number of steps. For small $K\tau$ ,
HeatSDF is a shell; larger diffusion produces a broader field. Details are in
Appendix~\ref{app:pde-details}.

\paragraph{Operators between surfaces and grids.}
A graph neural operator lifts features $f(p_j)$ at irregular surface samples
to grid queries $x_i$ using a learned kernel~\citep{li2020gno}. The mollified
version uses
\begin{equation}
 (\mathcal K_\theta f)(x_i)=\sum_j
 \rho\!\left(\frac{\|x_i-p_j\|_2}{r}\right)
 \widetilde\kappa_\theta(x_i,p_j)f(p_j)\,\omega_j,
 \label{eq:prelim-mgno}
\end{equation}
where $\omega_j$ are quadrature weights and the smooth mollifier $\rho$
vanishes outside radius $r$. Smoothly vanishing kernel contributions avoid
abrupt changes when moving samples cross a neighborhood boundary
\citep{lin2025mgno}. After lifting, FNO layers combine pointwise maps and
learned mixing of retained Fourier modes on the regular grid
\citep{li2021fno}. We use this mGNO--FNO composition for surface-to-shell
prediction; the downstream physics operator only requires a compatible
differentiable field-to-field map.

\section{Methodology}
\providecommand{\cM}{\mathcal{M}}
\providecommand{\cH}{\mathcal{H}}
\providecommand{\cG}{\mathcal{G}}
\providecommand{\cS}{\mathcal{S}}
\providecommand{\cR}{\mathcal{R}}
\providecommand{\bR}{\mathbb{R}}

\paragraph{Pipeline overview.}
We decompose the forward model into two composable neural operators.
Our mesh-to-shell operator (M2S-NO) $\cH_\theta$, implemented with mGNO--FNO, maps
oriented surface samples to a volumetric shell, such as TruncatedSDF or
HeatSDF with small diffusion time. A problem-specific PDE neural operator
(PDE-NO) $\cG_\theta$ maps
this field and the physical inputs to the desired solution. The second stage
is architecture-agnostic, provided its input and output spaces are compatible
and its input derivatives are available. A scalar objective $\mathcal J$,
such as peak displacement or integrated sensor response, supplies gradients
through both operators to the mesh. All learned weights are frozen during
inverse optimization, including those obtained by joint fine-tuning.
Algorithm~\ref{alg:inverse-shape} summarizes the procedure;
Appendix~\ref{app:topology-update} details the optional topology block.

\begin{algorithm}[t]
\caption{M2S-NO-based inverse shape optimization}
\label{alg:inverse-shape}
\begin{algorithmic}[1]
\Require Initial mesh $(V_0,F_0)$; frozen operators $\cH_\theta$, $\cG_\theta$;
objective $\mathcal J$; geometry losses $\mathcal L_{\mathrm{geo}}^{\mathcal M}$,
$\mathcal L_{\mathrm{geo}}^{\Phi}$ and weight $\lambda_{\mathrm{geo}}$;
step sizes $\eta_t$, $\eta_t^\phi$; topology and remeshing schedules
$\mathcal T_{\mathrm{topo}}$, $\mathcal T_R$

\For{$t=1,\ldots,T$}
 \State $(P_{t-1},N_{t-1})\leftarrow\operatorname{OrientedSurfaceSample}(V_{t-1},F_{t-1})$
 \State $\mathcal L_{\mathrm{design}}^{\mathcal M}\leftarrow
   \mathcal J\!\left(\cG_\theta\!\left(\cH_\theta(P_{t-1},N_{t-1})\right)\right)$
 \State $\mathcal L_{\mathrm{total}}^{\mathcal M}\leftarrow
   \mathcal L_{\mathrm{design}}^{\mathcal M}+
   \lambda_{\mathrm{geo}}\mathcal L_{\mathrm{geo}}^{\mathcal M}(V_{t-1},F_{t-1})$
 \State $V_t\leftarrow V_{t-1}-\eta_tP_F
   \nabla_{V_{t-1}}\mathcal L_{\mathrm{total}}^{\mathcal M}$, with $P_F$ from~\eqref{eq:mesh-preconditioner}
 \State Hold prescribed interface vertices at their fixed positions
 \State $F_t\leftarrow F_{t-1}$
 \If{$t\in\mathcal T_{\mathrm{topo}}$}
  \State $(V_t,F_t)\leftarrow
    \Call{TopologyUpdate}{V_t,F_t,\eta_t^\phi,\lambda_{\mathrm{geo}},\mathcal J,\cH_\theta,\cG_\theta,\mathcal L_{\mathrm{geo}}^{\Phi}}$
 \EndIf
 \If{$t\in\mathcal T_R$}
   \State $(V_t,F_t)\leftarrow\cR_{\mathrm{BK}}(V_t,F_t)$
 \EndIf
\EndFor
\end{algorithmic}
\end{algorithm}

\paragraph{Differentiable surface samples.}\label{par:surface-sampling}
Let $\cM=(V,F)$ be an oriented triangular mesh, with vertex positions
$V\in\bR^{n\times 3}$ and face indices
$F\in\{1,\ldots,n\}^{m\times 3}$.  Let $S$ be the number of oriented
surface samples.  For a fixed sampling state $\xi$, the differentiable sampler
returns oriented points and normals
$(P(V,F;\xi),N(V,F;\xi))\in\bR^{S\times3}\times\bR^{S\times3}$.
Faces are sampled according to surface area, and the selected faces and local
sample coordinates are held fixed during each gradient step.  Each point is
obtained by barycentric interpolation of its face vertices, and each normal by
interpolating and renormalizing the corresponding vertex normals.  Hence,
for fixed $\xi$, $(P,N)$ is differentiable with respect to $V$ away from
degenerate triangles; $F$ determines the connectivity and sampling support.
 Sampling and coordinate-normalization details are given in
Appendix~\ref{app:sampling-normalization}.

\paragraph{Forward surrogate models.}\label{par:forward-surrogates}
Let $\Phi$ denote the space of volumetric shell fields and let $\mathcal Y$
denote the space of physical output fields. M2S-NO maps oriented surface
samples to a shell representation,
\begin{equation}
 \cH_\theta:\bR^{S\times3}\times\bR^{S\times3}\to\Phi,
 \qquad \widehat\phi=\cH_\theta(P,N),
 \label{eq:forward-m2s}
\end{equation}
while the problem-specific PDE-NO maps the shell, together with fixed PDE coefficients and
boundary conditions omitted from the notation, to the physical field,
\begin{equation}
 \cG_\theta:\Phi\to\mathcal Y,
 \qquad \widehat y=\cG_\theta(\widehat\phi).
 \label{eq:forward-pde}
\end{equation}
The design objective $\mathcal J:\mathcal Y\to\bR$ is evaluated on
$\widehat y$.  During inverse optimization, both operators are frozen, so
gradients of the composite objective propagate through $\cG_\theta$ and
$\cH_\theta$ to the mesh vertices.

\paragraph{Geometry constraints.}\label{par:geometry-constraints}
Prescribed interface vertices are held fixed during mesh updates. The
remaining geometry loss uses differentiable soft penalties for constraints such as
volume, package-box containment, and thickness.  We use
$\mathcal L_{\mathrm{geo}}^{\mathcal M}(V,F)$ for the mesh-domain loss and
$\mathcal L_{\mathrm{geo}}^{\Phi}(\phi)$ for its shell-domain counterpart;
their numerical implementations may differ.  A collision penalty is included
only in $\mathcal L_{\mathrm{geo}}^{\mathcal M}$, so it affects mesh-space
updates but not the shell-space gradient step.  Any extracted shell candidate
is reevaluated with the full mesh-domain total loss, including this penalty,
before acceptance. Soft penalties encourage feasibility but do not enforce
exact equalities or inequalities. The mesh-domain collision, thickness, and
volume terms are defined in Appendix~\ref{app:geometry-losses}.

\paragraph{Preconditioned mesh update.}\label{par:preconditioner}
For the current mesh, let $n$ denote its current number of vertices, let
$L_F:=\operatorname{Lap}(V,F)\in\bR^{n\times n}$ denote its uniform mesh
Laplacian, and let $I_n$ be the $n\times n$ identity.  For
$\lambda_P>0$, define the smoothing matrix and induced preconditioner
\begin{equation}
 M_F=I_n+\lambda_P L_F,
 \qquad
 P_F=M_F^{-1}M_F^{-T}.
 \label{eq:mesh-preconditioner}
\end{equation}
The preconditioned update is
\begin{equation}
 V\leftarrow V-\eta_tP_F
 \nabla_V\mathcal L_{\mathrm{total}}^{\mathcal M}.
 \label{eq:preconditioned-update}
\end{equation}
Applying the inverse Laplacian factors
spreads localized gradients over neighboring vertices and suppresses
high-frequency deformation~\citep{nicolet2021large}.
The differential-coordinate implementation used to apply this preconditioner
is described in Appendix~\ref{app:inverse-details}. For constrained regions,
the fixed-vertex condition is applied to the resulting update: zeroing only
the raw gradient would not prevent the preconditioner from moving a fixed
vertex through its coupling to neighboring vertices.

\paragraph{Remeshing and topology change.}\label{par:topology-change}
Remeshing improves triangle quality and redistributes surface samples while
preserving topology. Because $\cH_\theta$ consumes oriented samples rather
than a fixed-length vertex vector, a change in vertex count does not require
retraining the surrogate. After remeshing, we rebuild the preconditioner and
optimizer state as described in Appendix~\ref{app:remeshing}. The shell-space
topology proposal is detailed separately in Appendix~\ref{app:topology-update}.
Protected interfaces must remain unchanged for their hard constraints to hold.
When $t\in\mathcal T_{\mathrm{topo}}$, we take a gradient step on the shell
field, extract its zero level set, and obtain a candidate mesh
$(V_\phi,F_\phi)$.  This extraction can change connectivity.  We accept the
candidate only if it changes topology and lowers the mesh-domain total loss;
otherwise, we keep $(V_t,F_t)$. This is a discrete proposal-and-acceptance
step, not backpropagation through a change of mesh topology. Unlike a
fixed-connectivity update, extraction does not automatically preserve
prescribed interfaces; we discuss this limitation in
Section~\ref{sec:limitations}.

\section{Experiments}
We evaluate M2S-NO on geometric matching and two physics-informed design
problems: jet-engine brackets~\citep{hong2025deepjeb} and neutron moderators
\citep{deng2026differentiable}.

\paragraph{Common setup for physics-informed optimization.}
We train M2S-NO on procedural shapes with varied geometry and topology, and
train a separate problem-specific PDE-NO to predict solution fields. For the
main experiments, we compose independently pretrained M2S-NO and PDE-NO models
and keep their weights fixed during inverse optimization. The appendix compares this
setting with joint fine-tuning using solution-field prediction loss.

To show the benefit of the shell interface, we compare four representations.
For each intermediate representation, we train two neural operators: M2S-NO
maps oriented surface samples to the intermediate field, and PDE-NO maps that
field and the physical inputs to the volumetric solution field. We use SDF as a
non-shell intermediate and TruncatedSDF and HeatSDF as shell intermediates. In
the \emph{Direct} baseline, we instead train a single neural operator that
takes surface samples together with physical inputs and directly predicts the
solution field, without an intermediate interface. The main text uses fixed
weights; the corresponding joint fine-tuning comparison is reported in
Appendix~\ref{app:finetune-comparison}. For each problem, \emph{Dataset best}
is the best design in its problem-specific dataset. Comparing against this
reference tests whether optimization can obtain designs that outperform those
already observed in the dataset. GANO is a representative
baseline for learned latent-space methods and uses a neural operator as its
forward surrogate~\citep{sun2026gano}. For neutron moderation, we additionally
compare with a traditional method based on a parametric representation.
Optimized physical performance is checked with independent numerical solvers.

\subsection{Category-Agnostic Shape-to-Shape Optimization}
\label{sec:shape-to-shape}

We generate a 4,000-shape procedural dataset using CSG operations over
primitives and analytic shape families, including spherical-harmonic surfaces,
cones, spheres, and cubes.  The dataset is designed to cover general geometry
rather than a fixed set of semantic object categories.

We train M2S-NO on this dataset and evaluate whether it provides useful shape
gradients between geometrically and topologically distinct objects.  Given a
source mesh and a target shell representation $\phi^*$, we use the identity
map in place of the PDE-NO in Algorithm~\ref{alg:inverse-shape}.  The optimization
objective is the discrepancy between the predicted and target shell fields:
\begin{equation}
 \cG_\theta=I_\Phi,
 \qquad
 \mathcal J(\cG_\theta(\cH_\theta(P,N)))=
 \|\cH_\theta(P,N)-\phi^*\|.
 \label{eq:shape-to-shape-objective}
\end{equation}

We compare M2S-NO with DeepSDF~\citep{park2019deepsdf}, an implicit
auto-decoder that represents each shape with an optimized latent code.  The
DeepSDF decoder is trained on the same procedural-shape dataset.  We first
optimize its latent code to reconstruct the initial sphere, which is included
in the training set, and then optimize the code toward the target SDF.

\begin{figure*}[t]
\centering
\begin{tikzpicture}
  \node[inner sep=0,anchor=south west] (img) at (0,0)
 {\includegraphics[width=0.92\textwidth]{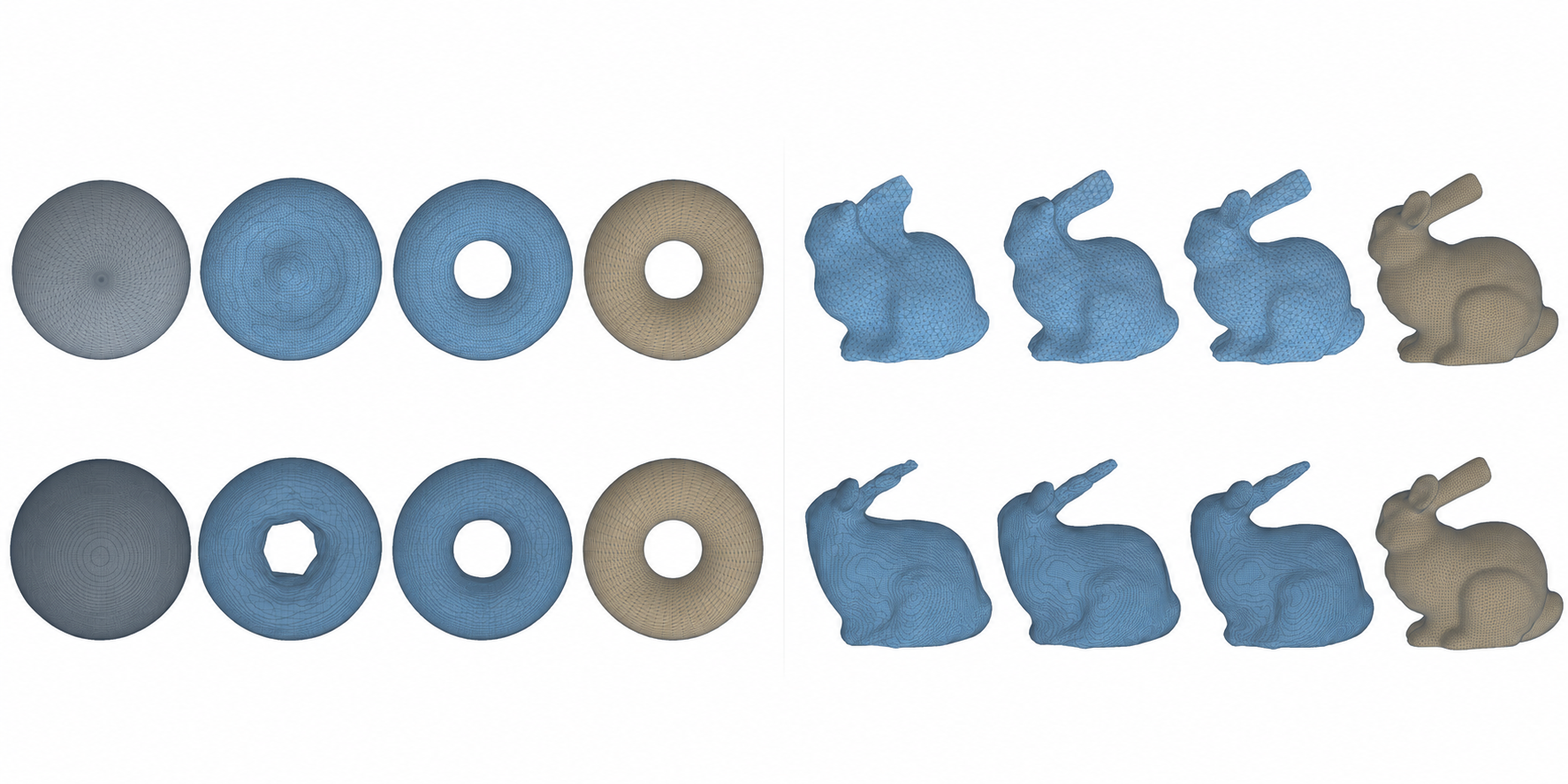}};
  \begin{scope}[x={($(img.south east)-(img.south west)$)},
y={($(img.north west)-(img.south west)$)}]
\node at (0.0625,0.88) {Source};
\node at (0.1875,0.88) {Step 50};
\node at (0.3125,0.88) {Step 375};
\node at (0.4375,0.88) {Target};
\node at (0.5625,0.88) {Step 50};
\node at (0.6875,0.88) {Step 100};
\node at (0.8125,0.88) {Step 375};
\node at (0.9375,0.88) {Target};
\node[rotate=90] at (-0.035,0.63) {M2S-NO (ours)};
\node[rotate=90] at (-0.035,0.275) {DeepSDF};
\end{scope}
\end{tikzpicture}
\caption{Category-agnostic shape matching: sphere-to-torus (left) and
sphere-to-bunny (right). Rows compare M2S-NO and DeepSDF trained on the same
procedural dataset. Both recover the torus topology; M2S-NO better resolves
the bunny's separated ears. Both tasks start from a sphere.}
\label{fig:shape-to-shape}
\end{figure*}
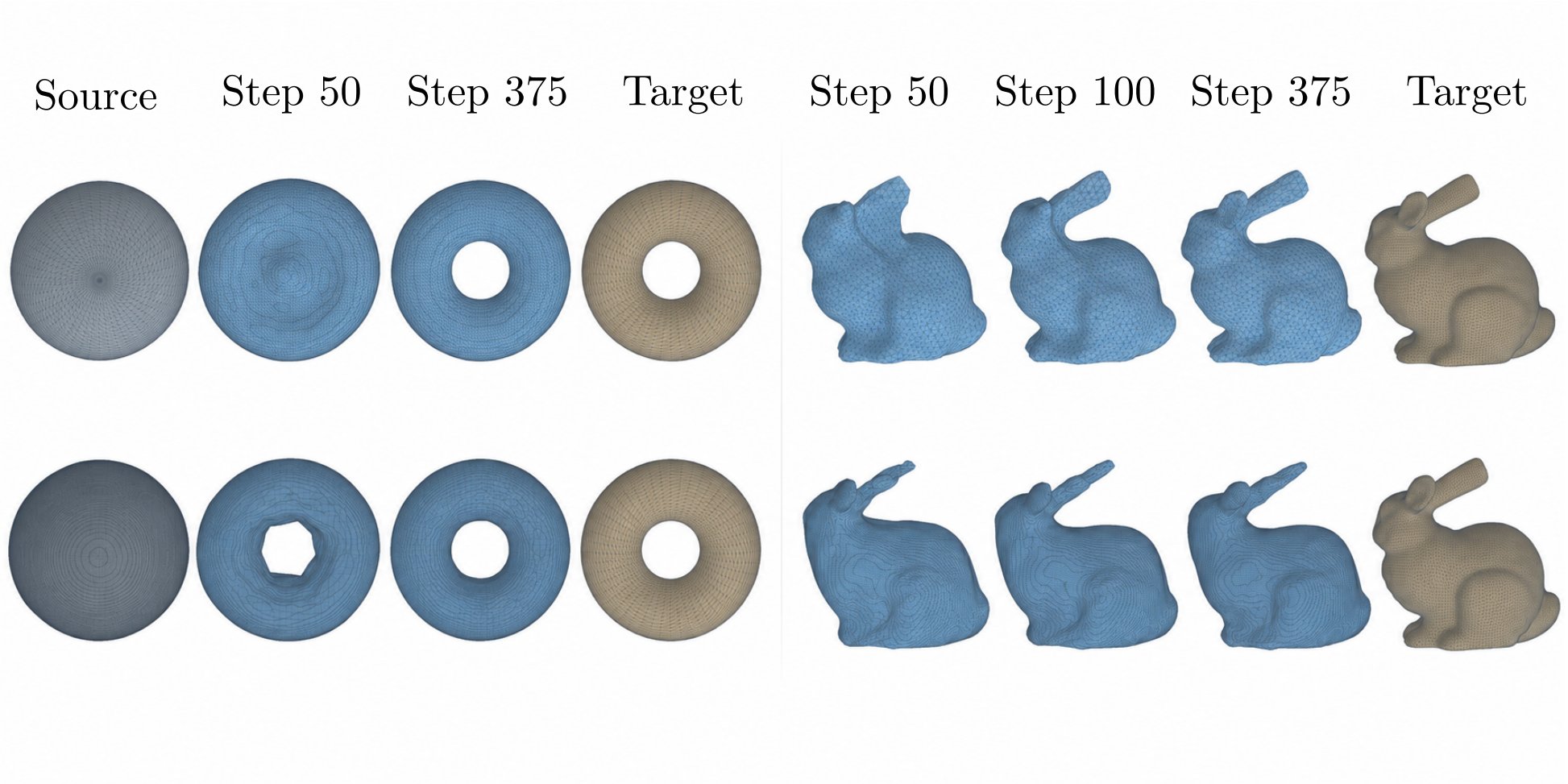

Figure~\ref{fig:shape-to-shape} shows the resulting optimization trajectories.
For the sphere-to-bunny task, DeepSDF recovers the coarse shape but does not
resolve the separated ears, consistent with limited decoder expressivity for
this target. M2S-NO deforms the source sphere toward the bunny
while preserving mesh connectivity throughout optimization. For the
sphere-to-torus task, M2S-NO uses the topology-update block to change the
surface genus and reaches the target topology.  DeepSDF also handles this
topology change successfully.

\subsection{DeepJEB Displacement Optimization}
\label{sec:deepjeb}

We evaluate our method on the DeepJEB synthetic jet-engine bracket dataset
\citep{hong2025deepjeb}, which contains 2,138 FEM-labeled bracket geometries.
The problem-specific PDE-NO predicts the displacement-resultant field
$u(x;\mathcal S)$ in the bracket volume under the prescribed upward load. We
keep the loaded and bolted interfaces fixed while optimizing the remaining
surface.

We solve the interface-constrained problem
\begin{equation}
 \begin{aligned}
  \min_{\mathcal S}\quad
  &\mathcal J(\mathcal S)=
\max_{x\in\Omega(\mathcal S)} u(x;\mathcal S) \\
  \text{s.t.}\quad
  &\Gamma_{\mathrm{load}}(\mathcal S)
=\Gamma_{\mathrm{load}}^*,
  \qquad
  \Gamma_{\mathrm{bolt}}(\mathcal S)
=\Gamma_{\mathrm{bolt}}^*.
 \end{aligned}
 \label{eq:deepjeb-objective}
\end{equation}
Here $\Gamma_{\mathrm{load}}^*$ and $\Gamma_{\mathrm{bolt}}^*$ are the shared,
prescribed interfaces. Their vertices are excluded from updates. We optimize
a normalized soft maximum and evaluate the true maximum with an independent
FEM solver; this objective measures peak deformation rather than stress or
failure.

We test both in-distribution and geometric extrapolation. The
\emph{in-distribution bracket} is the dataset bracket with sample ID 101--428;
the \emph{arch-augmented OOD bracket} is formed by adjoining a CAD arch to it.
Both retain the same fixed interfaces.

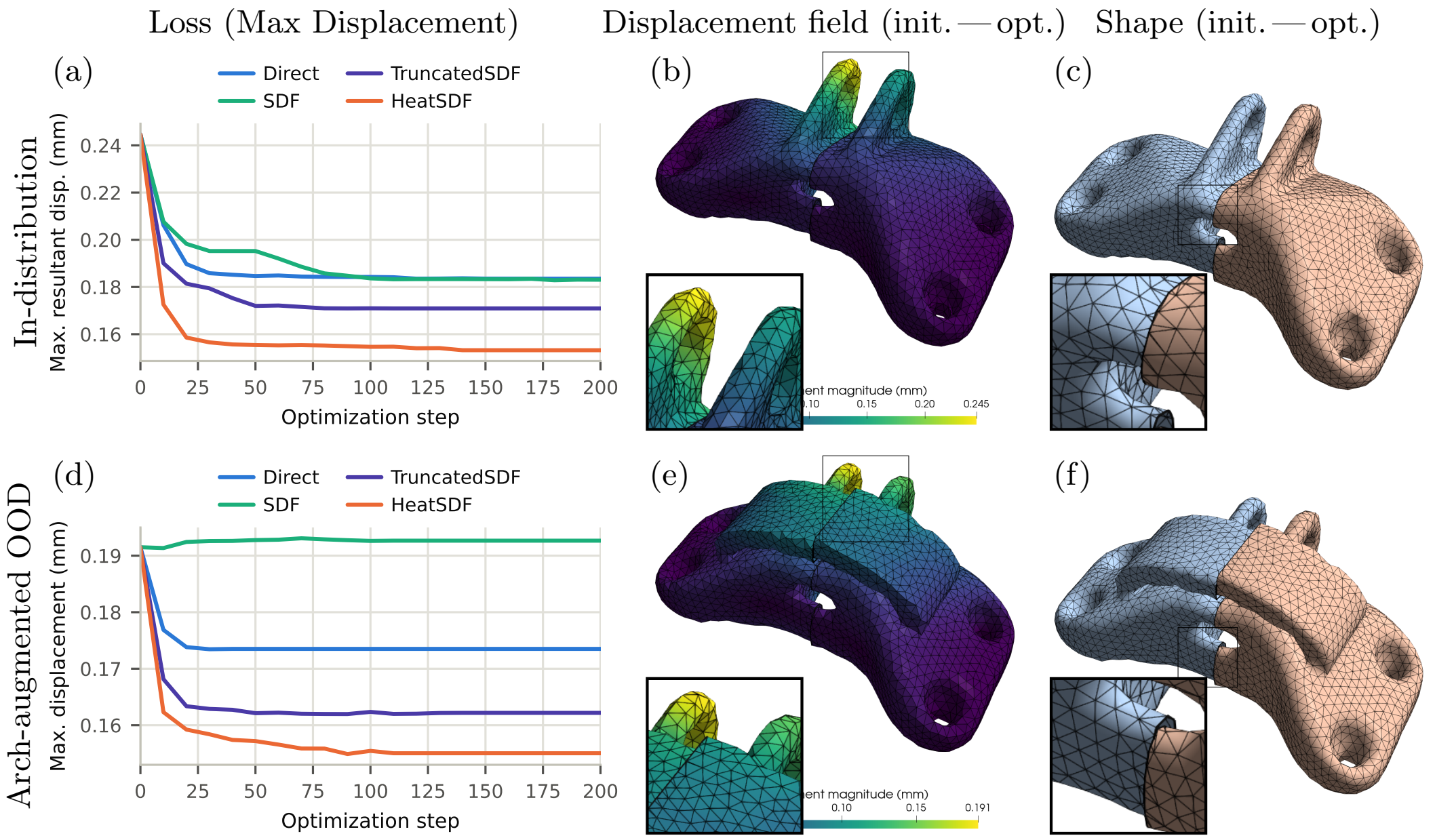
\begin{figure}[t]
  \centering
  \pgfmathsetlengthmacro{\cellsize}{0.27\textwidth}  
  \pgfmathsetlengthmacro{\losswidth}{1.5*\cellsize}  
  \pgfmathsetlengthmacro{\cellgap}{0.012\textwidth}  
  \pgfmathsetlengthmacro{\zoomsize}{0.40*\cellsize}
  \pgfmathsetlengthmacro{\zoominset}{0.015*\cellsize}
  \begin{tikzpicture}[
      img/.style={inner sep=0pt, outer sep=0pt, anchor=north west},
      head/.style={font=\small, anchor=south, inner sep=2pt},
      rowlab/.style={font=\small, rotate=90, anchor=south, inner sep=2pt},
      spy using outlines={rectangle, draw=black, line width=0.6pt, magnification=2.6},
    ]
    \def\colA{0}
    \pgfmathsetlengthmacro{\colB}{\losswidth+\cellgap}
    \pgfmathsetlengthmacro{\colC}{\losswidth+\cellsize+2*\cellgap}
    \def\rowA{0}
    \pgfmathsetlengthmacro{\rowB}{-\cellsize-\cellgap}
    \node[img] (AA) at (\colA,\rowA) {\includegraphics[width=\losswidth,height=\cellsize]{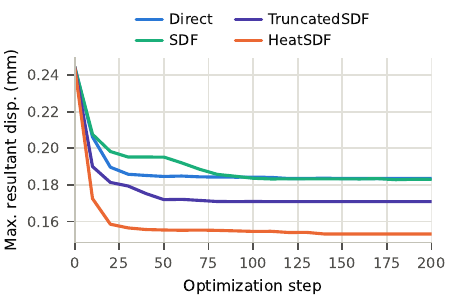}};
    \node[img] (AB) at (\colB,\rowA) {\includegraphics[width=\cellsize,height=\cellsize]{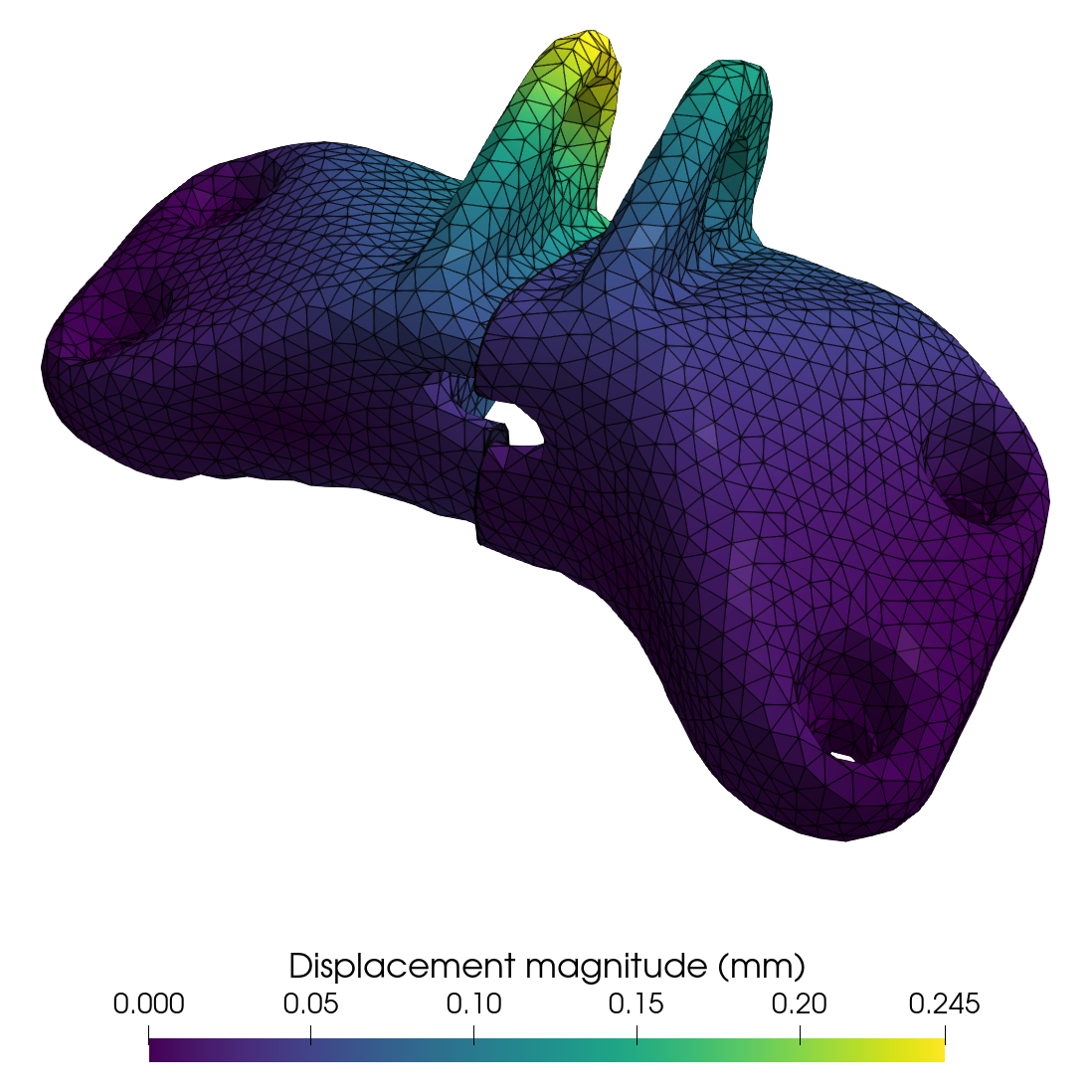}};
    \node[img] (AC) at (\colC,\rowA) {\includegraphics[width=\cellsize,height=\cellsize]{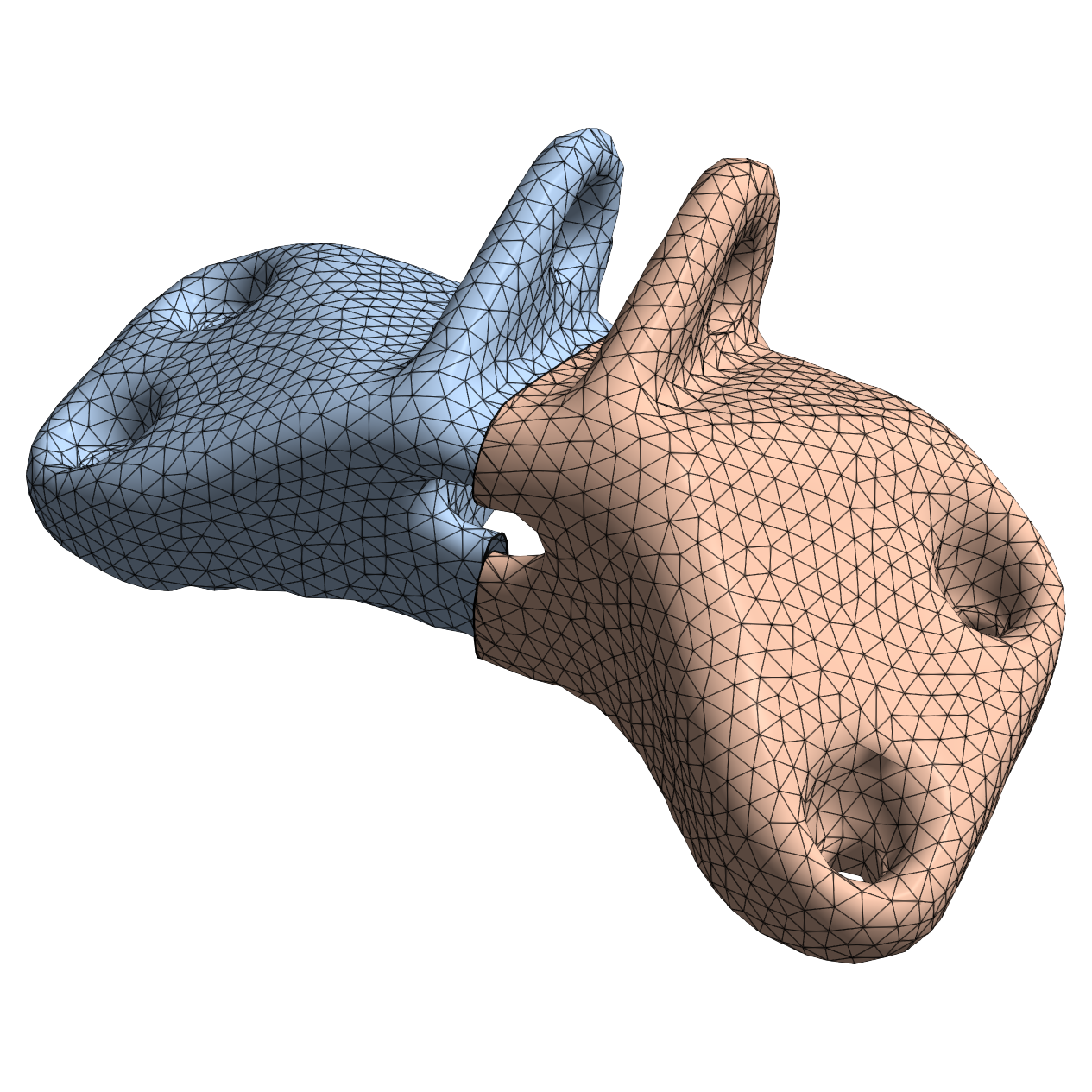}};
    \node[img] (BA) at (\colA,\rowB) {\includegraphics[width=\losswidth,height=\cellsize]{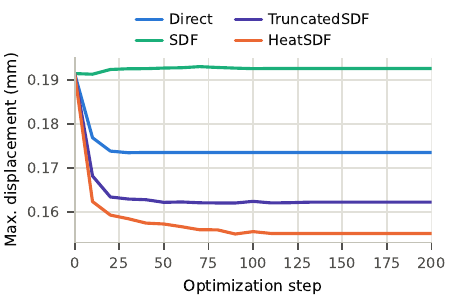}};
    \node[img] (BB) at (\colB,\rowB) {\includegraphics[width=\cellsize,height=\cellsize]{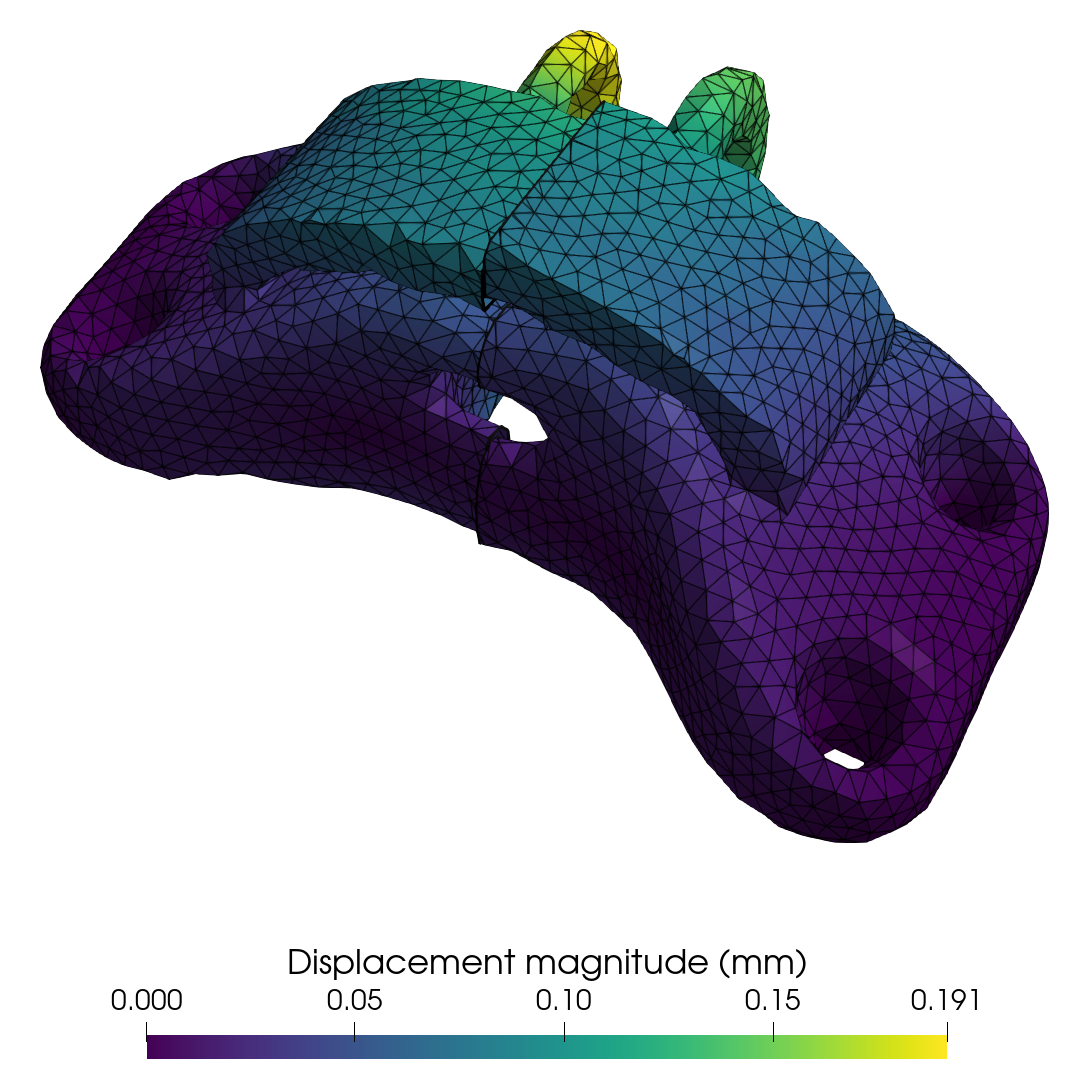}};
    \node[img] (BC) at (\colC,\rowB) {\includegraphics[width=\cellsize,height=\cellsize]{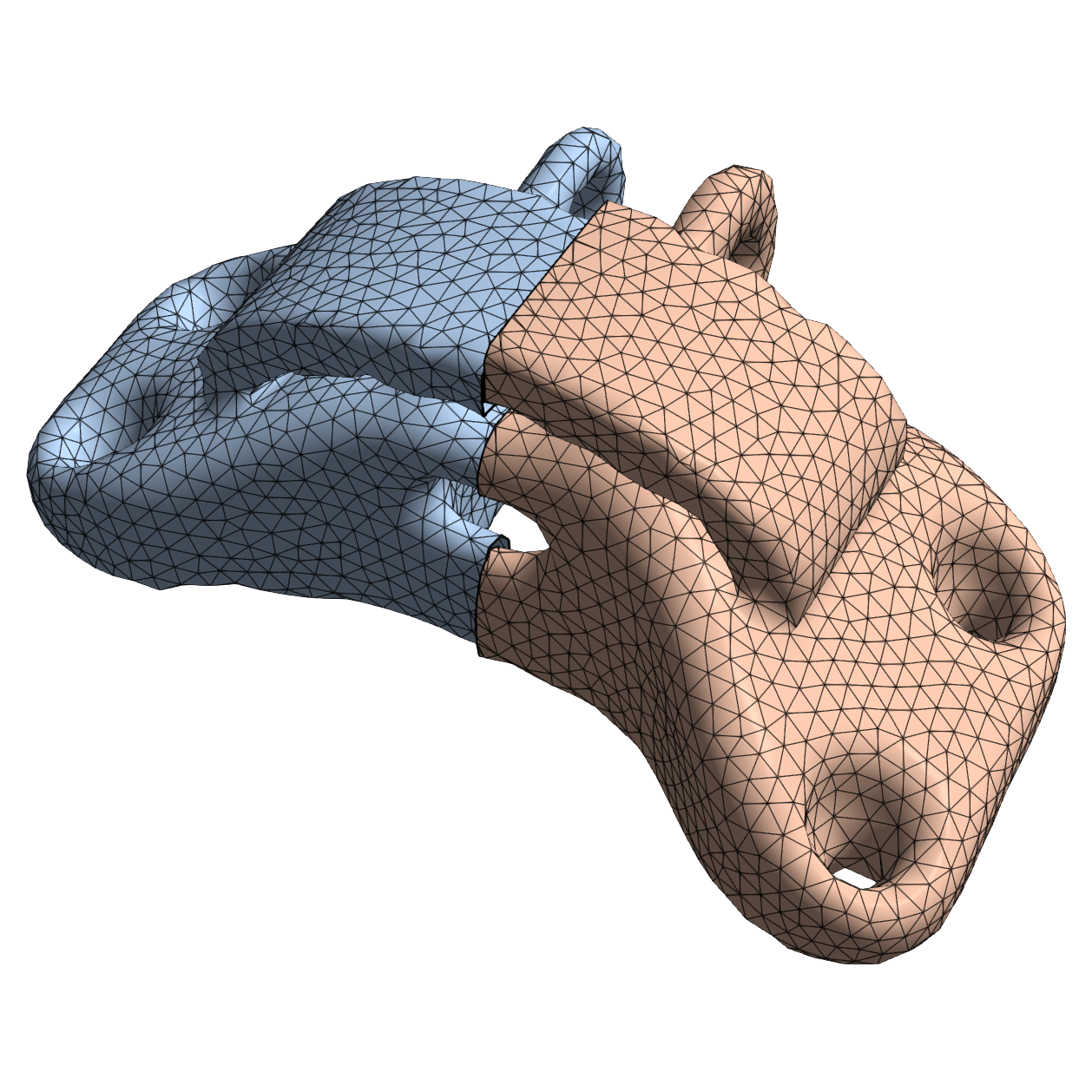}};
    \spy[size=\zoomsize, magnification=1.8] on ($(AB.north west)+(0.58*\cellsize,-0.12*\cellsize)$)
      in node[anchor=south west] at ($(AB.south west)+(\zoominset,\zoominset)$);
    \spy[size=\zoomsize, magnification=1.8] on ($(BB.north west)+(0.58*\cellsize,-0.12*\cellsize)$)
      in node[anchor=south west] at ($(BB.south west)+(\zoominset,\zoominset)$);
    \spy[size=\zoomsize] on ($(AC.north west)+(0.42*\cellsize,-0.43*\cellsize)$)
      in node[anchor=south west] at ($(AC.south west)+(\zoominset,\zoominset)$);
    \spy[size=\zoomsize] on ($(BC.north west)+(0.42*\cellsize,-0.53*\cellsize)$)
      in node[anchor=south west] at ($(BC.south west)+(\zoominset,\zoominset)$);
    \node[head] at (AA.north) {Loss (Max Displacement)};
    \node[head] at (AB.north) {Displacement field (init.\,|\,opt.)};
    \node[head] at (AC.north) {Shape (init.\,|\,opt.)};
    \node[rowlab] at (AA.west) {In-distribution};
    \node[rowlab] at (BA.west) {Arch-augmented OOD};
    \foreach \n/\l in {AA/a, AB/b, AC/c, BA/d, BB/e, BC/f}
      \node[font=\small, anchor=north west, inner sep=2pt] at (\n.north west) {(\l)};
  \end{tikzpicture}
  \caption{DeepJEB bracket optimization for minimizing maximum displacement.
  The upper and lower rows use the in-distribution bracket (sample ID 101--428)
  and the arch-augmented OOD bracket, respectively. In the first column (a,d),
  the FEM-evaluated maximum displacement curves compare shell representations (HeatSDF, TruncatedSDF) and baselines (Direct, SDF); Shell representations outperform the baselines. The second column (b,e)
  visualizes the displacement fields for the HeatSDF optimization, formed by
  joining the initial shape on the left and the optimized shape on the right;
  the insets enlarge the load-interface region. The third column (c,f) shows
  the corresponding initial--optimized shape composites, with insets at the
  join.}
 \label{fig:deepjeb-geometries}
\end{figure}
Figure~\ref{fig:deepjeb-geometries} uses two initializations to evaluate the
optimization. In the loss plots (a,d), Direct, SDF, TruncatedSDF, and HeatSDF
are compared; both shell representations outperform the baselines, with
HeatSDF achieving the lowest maximum displacement. In the field composites
(b,e), the left and right halves show the initial and optimized fields,
respectively. The load-interface insets show that the optimized field is darker
than the initial field, indicating lower displacement near the load interface.
The shape composites (c,f) show the corresponding geometric change: the
initial and optimized brackets are visibly misaligned in the vertical
direction, indicating a clear vertical change.

\begin{table}[t]
\centering
\small
\caption{FEM-evaluated maximum displacement resultant (mm; lower is better).
Each row is one starting shape and is compared horizontally. \emph{Dataset
best} is an independent reference; GANO and M2S-NO (ours) are optimized from
that starting shape. GANO is unavailable for the arch-augmented OOD case.}
\label{tab:deepjeb-fem-results}
\begin{tabular}{lcccc}
\toprule
\multicolumn{2}{c}{Starting point} & \multicolumn{1}{c}{Independent reference}
& \multicolumn{2}{c}{Optimization results} \\
\cmidrule(lr){1-2}\cmidrule(lr){3-3}\cmidrule(lr){4-5}
Starting shape & Initial value & Dataset best & GANO & M2S-NO (ours) \\
\midrule
\shortstack[l]{In-distribution} & 0.2446 & 0.1788 & 0.1884 & \textbf{0.1532} \\
\shortstack[l]{Arch-augmented OOD} & 0.1915 & 0.1788 & --     & \textbf{0.1549} \\
\bottomrule
\end{tabular}
\end{table}

The independent FEM results in Table~\ref{tab:deepjeb-fem-results} show that
M2S-NO obtains the best design for both starting shapes, using fixed pretrained
weights. For the in-distribution bracket, it reduces maximum displacement from
0.2446 to 0.1532, a $37.4\%$ reduction; the result is $14.3\%$ below the
problem-specific Dataset best and $18.7\%$ below GANO. On the
arch-augmented OOD bracket, it reduces displacement from 0.1915 to 0.1549, a
$19.1\%$ reduction, and still improves on Dataset best by $13.4\%$. Thus,
M2S-NO improves over both learned and dataset references even when the
starting geometry includes an unseen arch. GANO's StableSDF auto-decoder cannot fit
the OOD composite, so no GANO trajectory is available for that starting shape
(Appendix~\ref{app:gano-stablesdf}).

\subsection{Neutron Moderator Optimization}
\label{sec:neutron-current}
Neutron moderators improve detector response by scattering fast neutrons into
the lower energy range to which the detector is more sensitive.  Because this
transport depends strongly on geometry, moderator design is a challenging
inverse problem~\citep{deng2026differentiable}.  Our experimental setup follows
the neutron flux monitor setting of \citet{wang2024neutronics}.
We optimize the moderator surface
$\mathcal S$ to maximize the incoming middle-energy-group current on the side
of a cylindrical sensor:
\begin{equation}
 \begin{aligned}
  \max_{\mathcal S}\quad
  &Q_M(\mathcal S)=\int_{\Gamma_{\mathrm{sensor}}}
    J_M^{\mathrm{in}}(z,\theta;\mathcal S)\,\mathrm dA \\
  \text{s.t.}\quad
  &\Omega_{\mathrm{sensor}}\subset\Omega(\mathcal S),
  \qquad t_{\min}\le\operatorname{thick}_{\mathcal S}(x)\le t_{\max}
  \quad \forall x\in\mathcal S, \\
  &\operatorname{Vol}(\Omega(\mathcal S))\le V_{\max}.
 \end{aligned}
 \label{eq:neutron-current-objective}
\end{equation}
Here $J_M^{\mathrm{in}}$ is the incoming middle-energy current density,
$\Omega_{\mathrm{sensor}}$ is the sensor cylinder, and $\Omega(\mathcal S)$ is
the region enclosed by the moderator.  The lower and upper thickness limits,
$t_{\min}$ and $t_{\max}$, together with the volume budget $V_{\max}$, encode
standard engineering constraints.

For this problem, we generate a constrained procedural dataset of $100{,}000$
moderator designs using CSG.  Unlike the category-agnostic dataset used to
train M2S-NO, these shapes are restricted to be connected and
to enclose the sensor cylinder. OpenMC~\citep{romano2015openmc} provides noisy transport labels for training
the PDE-NO to predict current density on a receiver grid over the cylindrical
sensor side. The problem-specific PDE-NO and its training protocol are based on
the Particle Transport Neural Operator (PTNO)~\citep{cao2026ptno}.
Dataset construction details are given in
Appendix~\ref{app:neutron-current}.  Optimized candidates are independently
evaluated with OpenMC.


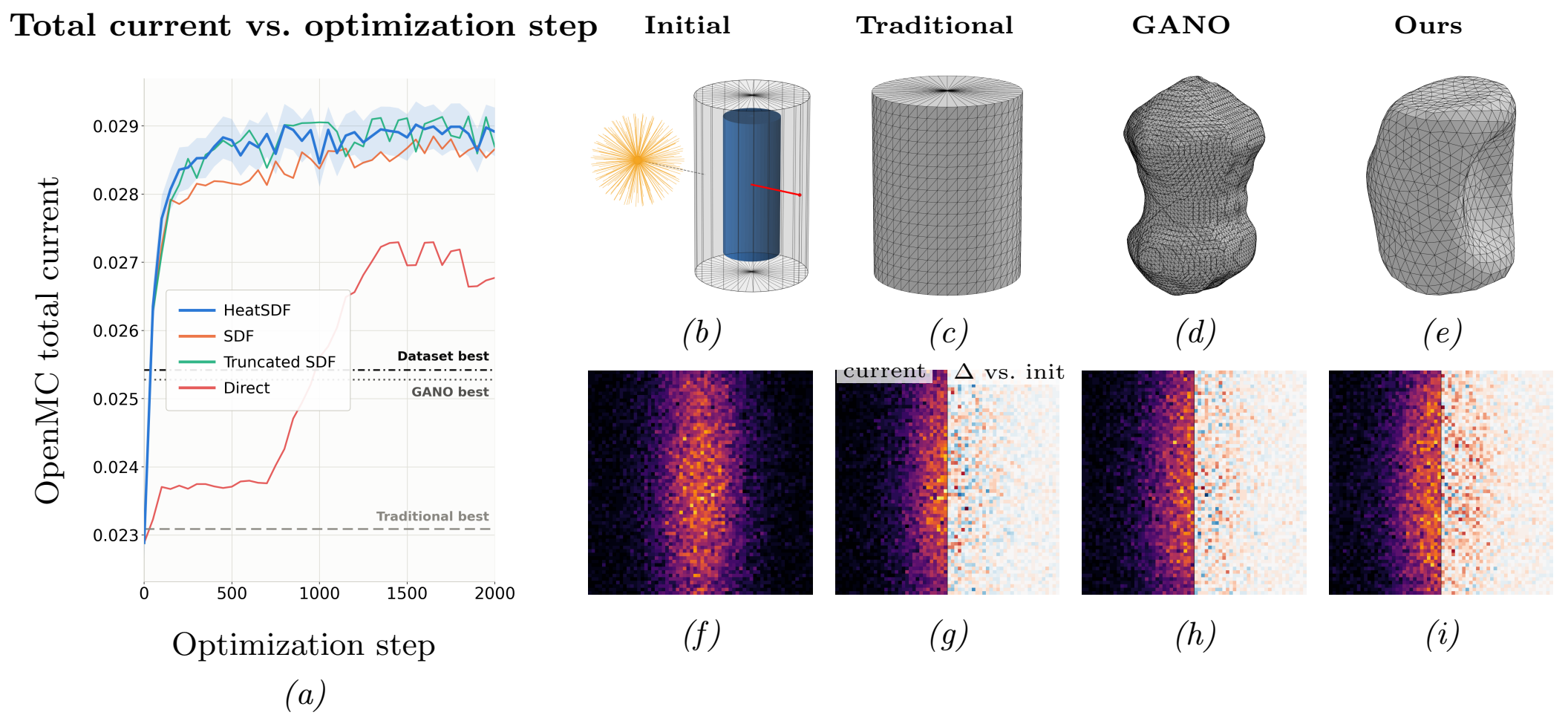
\begin{figure}[t]
\centering


\definecolor{cHeatSDF}{RGB}{42,120,214}
\definecolor{cSDF}{RGB}{235,104,52}
\definecolor{cTruncSDF}{RGB}{27,175,122}
\definecolor{cDirect}{RGB}{227,73,72}
\definecolor{cTraditionalBest}{RGB}{137,135,129}
\definecolor{cGANOBest}{RGB}{82,81,78}
\definecolor{cDatasetBest}{RGB}{11,11,11}

\newlength{\curvewidth} \setlength{\curvewidth}{0.300\textwidth} 
\newlength{\colgap}     \setlength{\colgap}{0.030\textwidth}     
\newlength{\colsep}     \setlength{\colsep}{0.016\textwidth}     
\newlength{\tinygap}    \setlength{\tinygap}{0.004\textwidth}    
\newlength{\ylabelgap}  \setlength{\ylabelgap}{0.003\textwidth}  
\newlength{\headgap}    \setlength{\headgap}{0.010\textwidth}    
\newlength{\rowgap}     \setlength{\rowgap}{0.033\textwidth}     
\newlength{\tagdrop}    \setlength{\tagdrop}{0.020\textwidth}    

\newlength{\fieldimg} 
\pgfmathsetlength{\fieldimg}{(\textwidth-\curvewidth-\colgap-3\colsep)/4}

\resizebox{\linewidth}{!}{%
\begin{tikzpicture}[
    label/.style={font=\small, inner sep=\tinygap},
    tag/.style={font=\small\itshape, inner sep=\tinygap},
]

\node[anchor=north west] (curve) at (0,0)    {\includegraphics[width=\curvewidth]{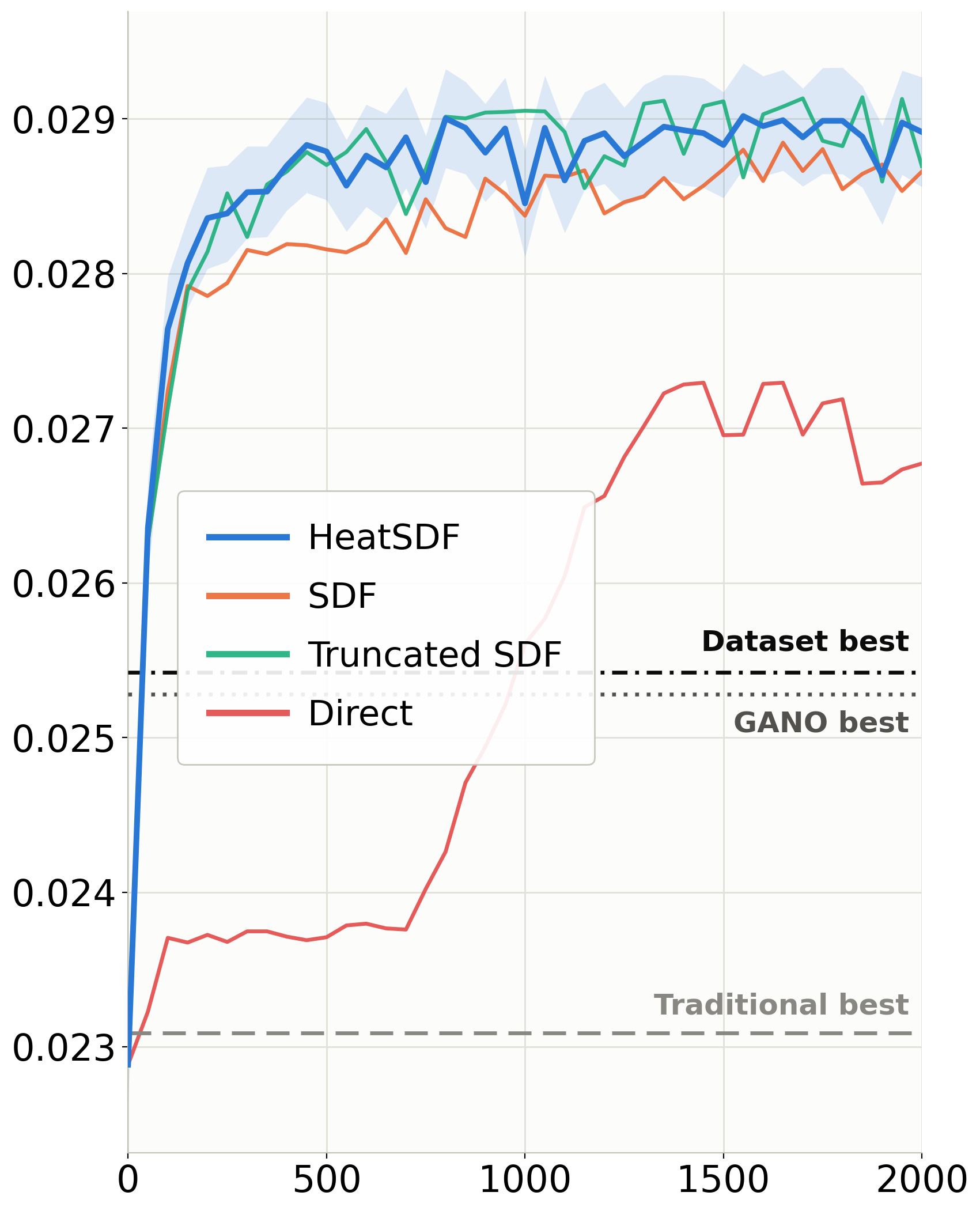}};

\node[font=\small\bfseries, above=\tinygap of curve.north]
    {Total current vs.\ optimization step};
\node[label, below=\tinygap of curve.south] (xlab) {\small Optimization step};
\node[label, rotate=90, anchor=south] at ([xshift=-\ylabelgap]curve.west)
    {\small OpenMC total current};
\node[tag, below=\tinygap of xlab.south] {(a)};

\coordinate (col1) at ([xshift=\colgap]curve.north east);
\coordinate (col2) at ($(col1)+(\fieldimg+\colsep,0)$);
\coordinate (col3) at ($(col1)+(2*\fieldimg+2*\colsep,0)$);
\coordinate (col4) at ($(col1)+(3*\fieldimg+3*\colsep,0)$);

\node[font=\scriptsize\bfseries, anchor=south] at ($(col1)+(0.5*\fieldimg,\headgap)$) {Initial};
\node[font=\scriptsize\bfseries, anchor=south] at ($(col2)+(0.5*\fieldimg,\headgap)$) {Traditional};
\node[font=\scriptsize\bfseries, anchor=south] at ($(col3)+(0.5*\fieldimg,\headgap)$) {GANO};
\node[font=\scriptsize\bfseries, anchor=south] at ($(col4)+(0.5*\fieldimg,\headgap)$) {Ours};

\node[anchor=north west] (rowA1) at (col1)
    {\includegraphics[width=\fieldimg]{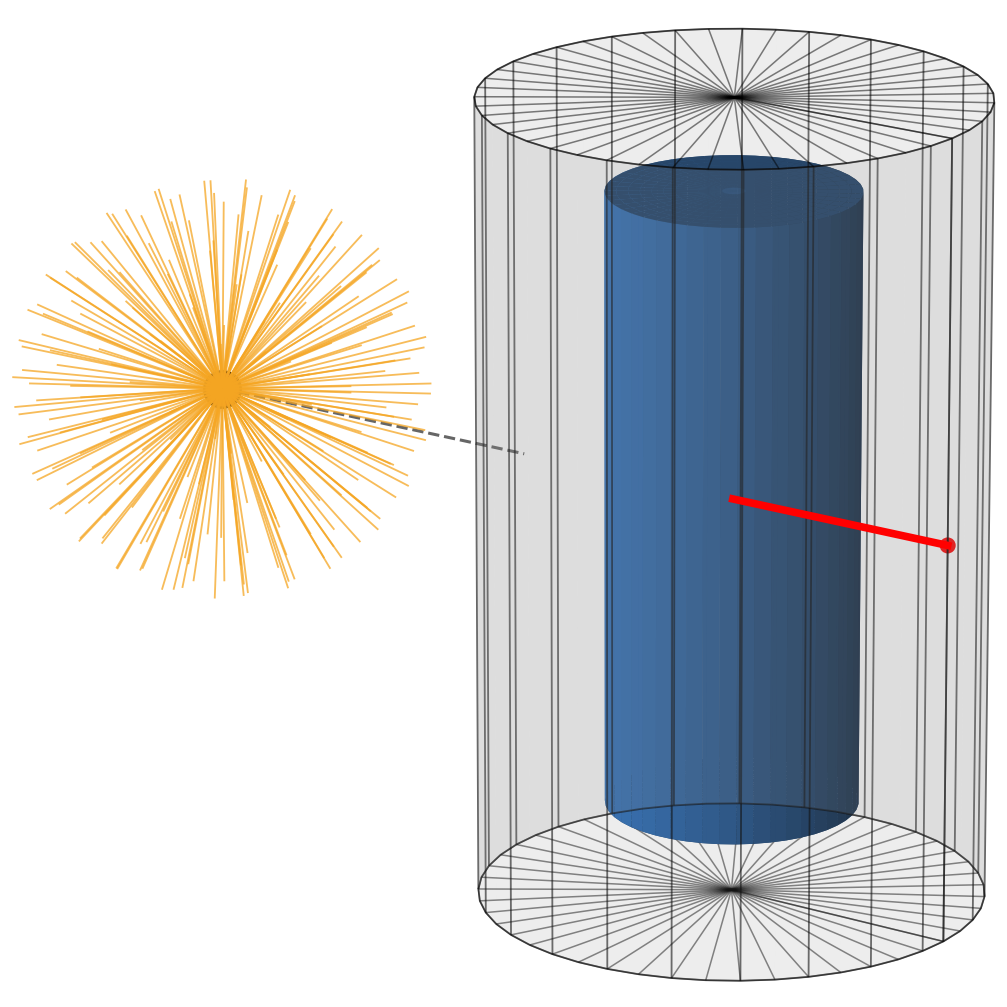}};
\node[anchor=north west] (rowA2) at (col2)
    {\includegraphics[width=\fieldimg]{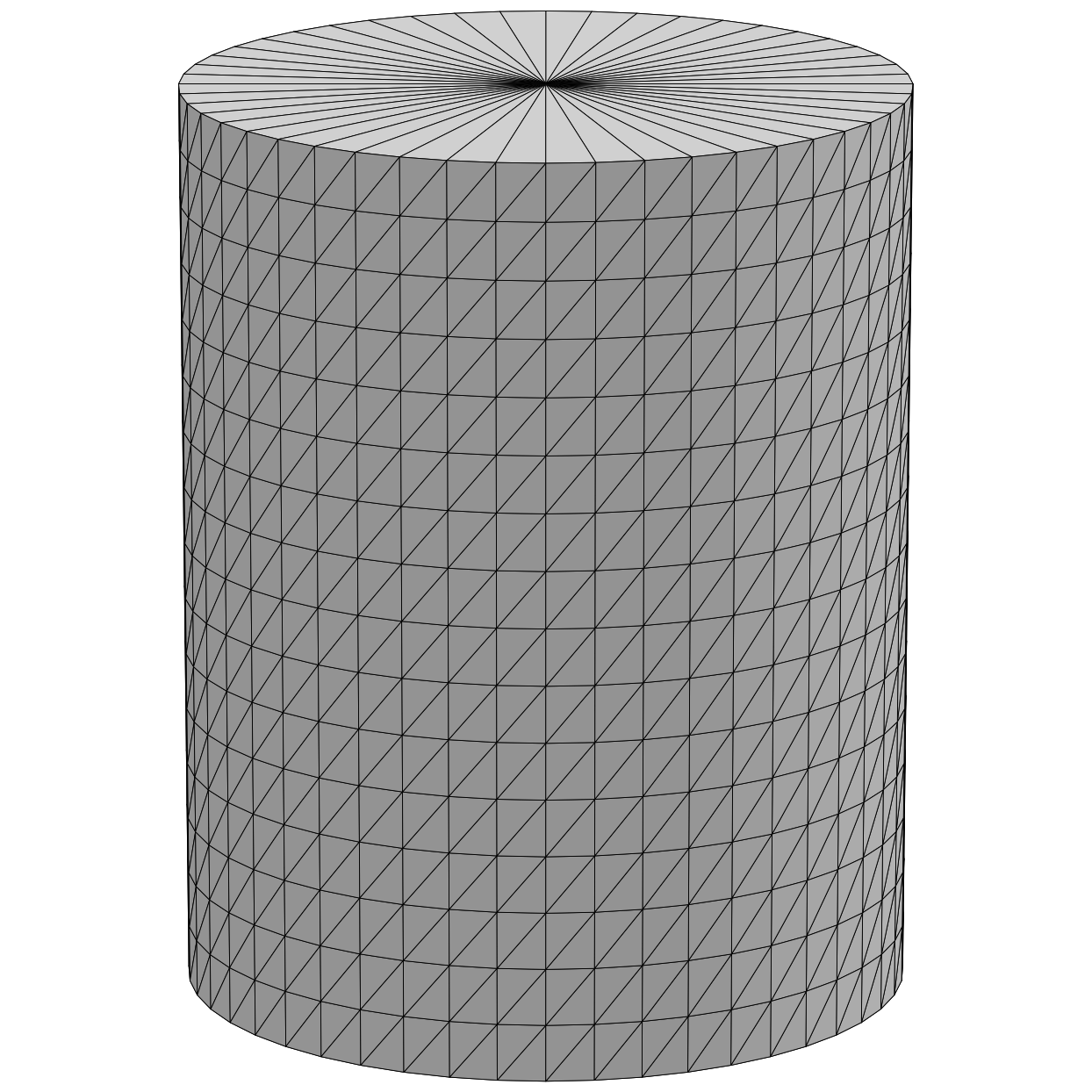}};
\node[anchor=north west] (rowA3) at (col3)
    {\includegraphics[width=\fieldimg]{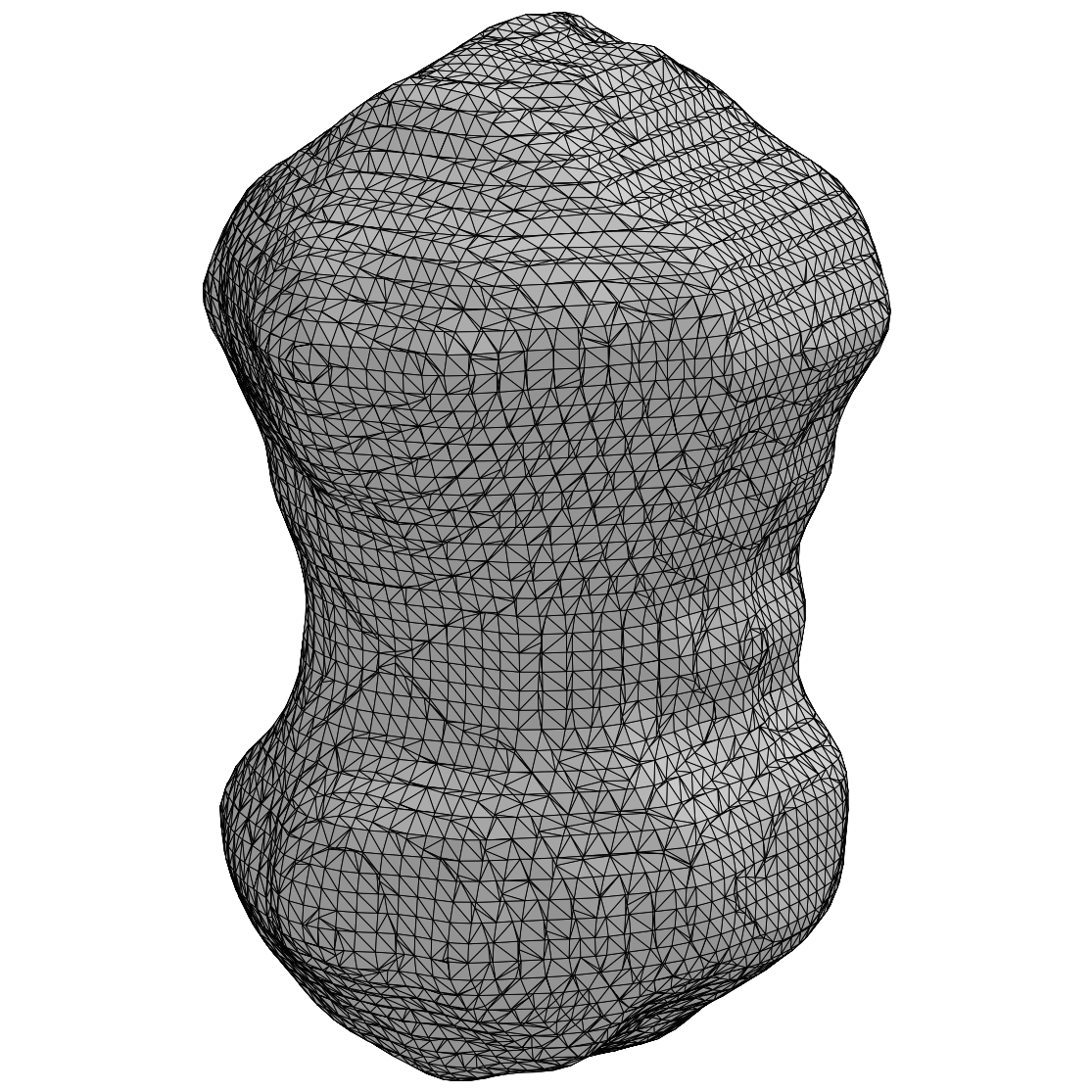}};
\node[anchor=north west] (rowA4) at (col4)
    {\includegraphics[width=\fieldimg]{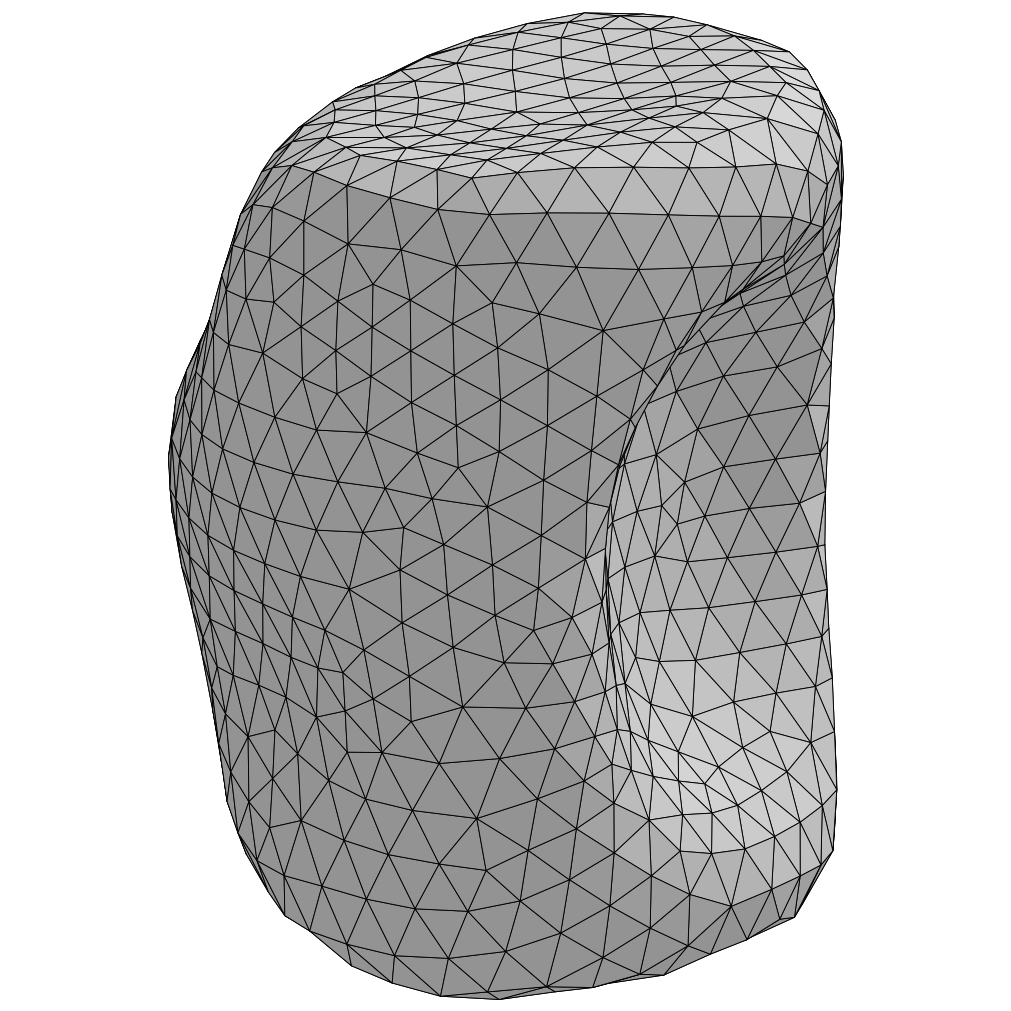}};

\coordinate (srcTop) at ($(rowA1.north west)!0.22!(rowA1.north east)$);
\coordinate (srcBot) at ($(rowA1.south west)!0.22!(rowA1.south east)$);

\node[anchor=north west] (rowB1) at ([yshift=-\rowgap]rowA1.south west)
    {\includegraphics[width=\fieldimg]{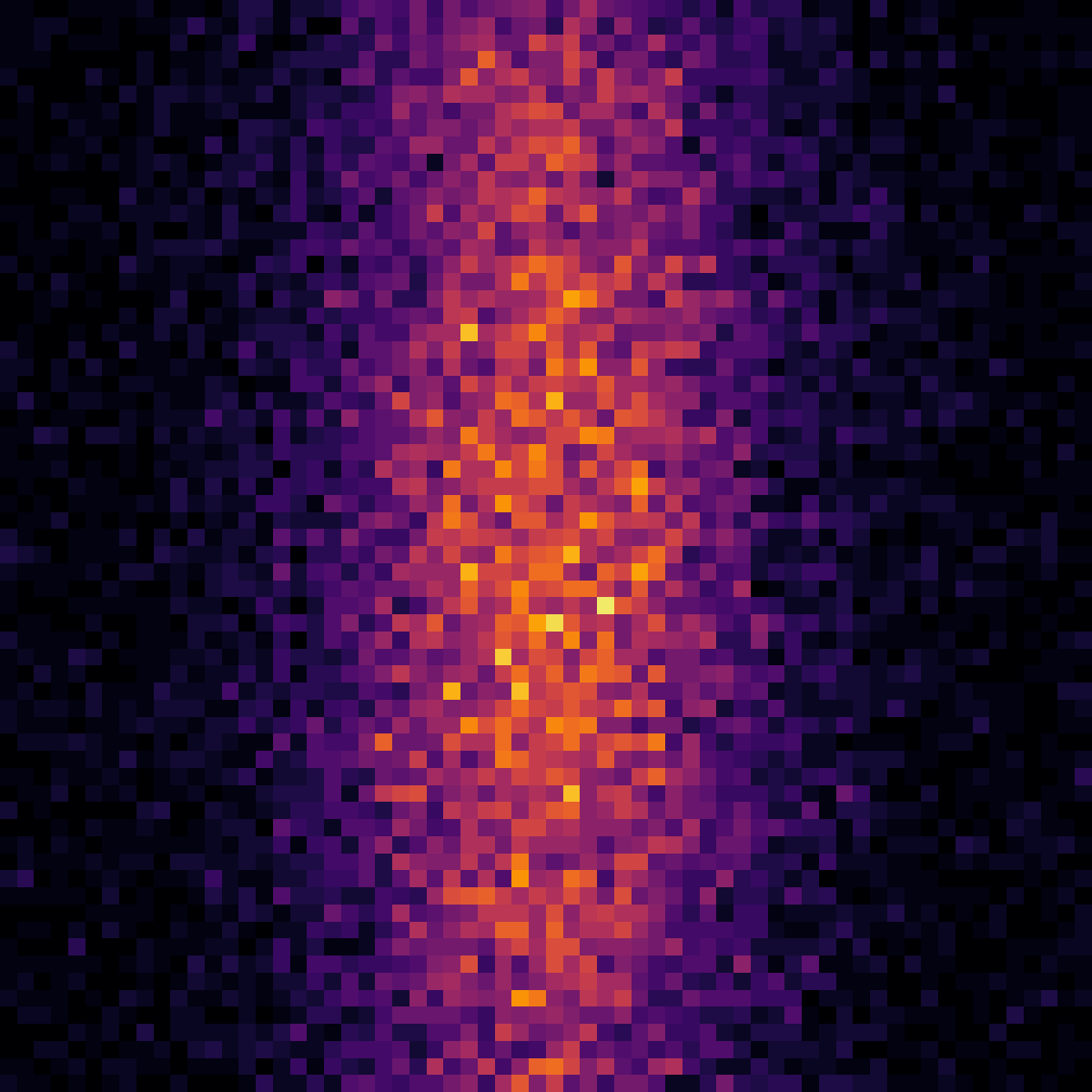}};
\node[anchor=north west] (rowB2) at ([yshift=-\rowgap]rowA2.south west)
    {\includegraphics[width=\fieldimg]{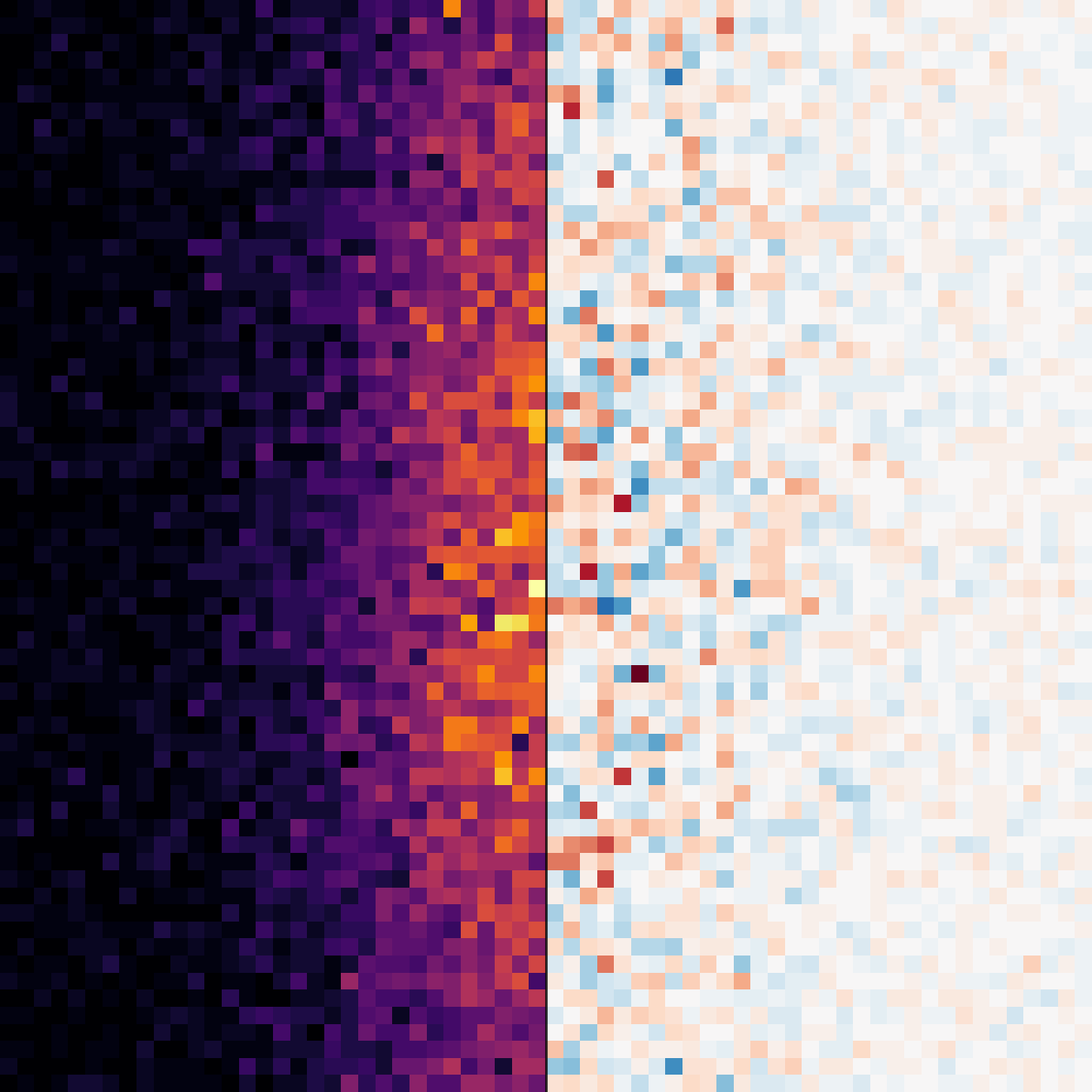}};
\node[anchor=north west] (rowB3) at ([yshift=-\rowgap]rowA3.south west)
    {\includegraphics[width=\fieldimg]{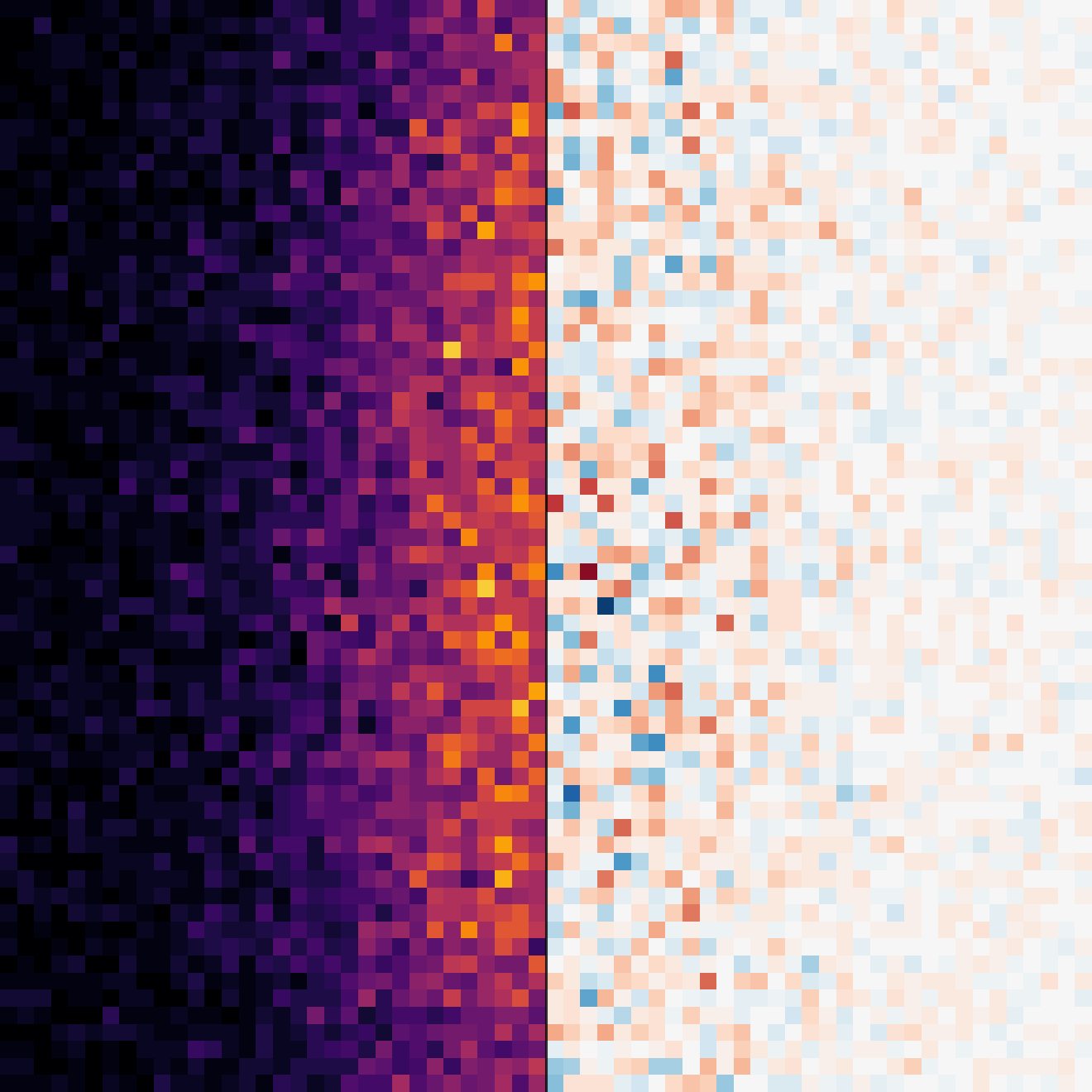}};
\node[anchor=north west] (rowB4) at ([yshift=-\rowgap]rowA4.south west)
    {\includegraphics[width=\fieldimg]{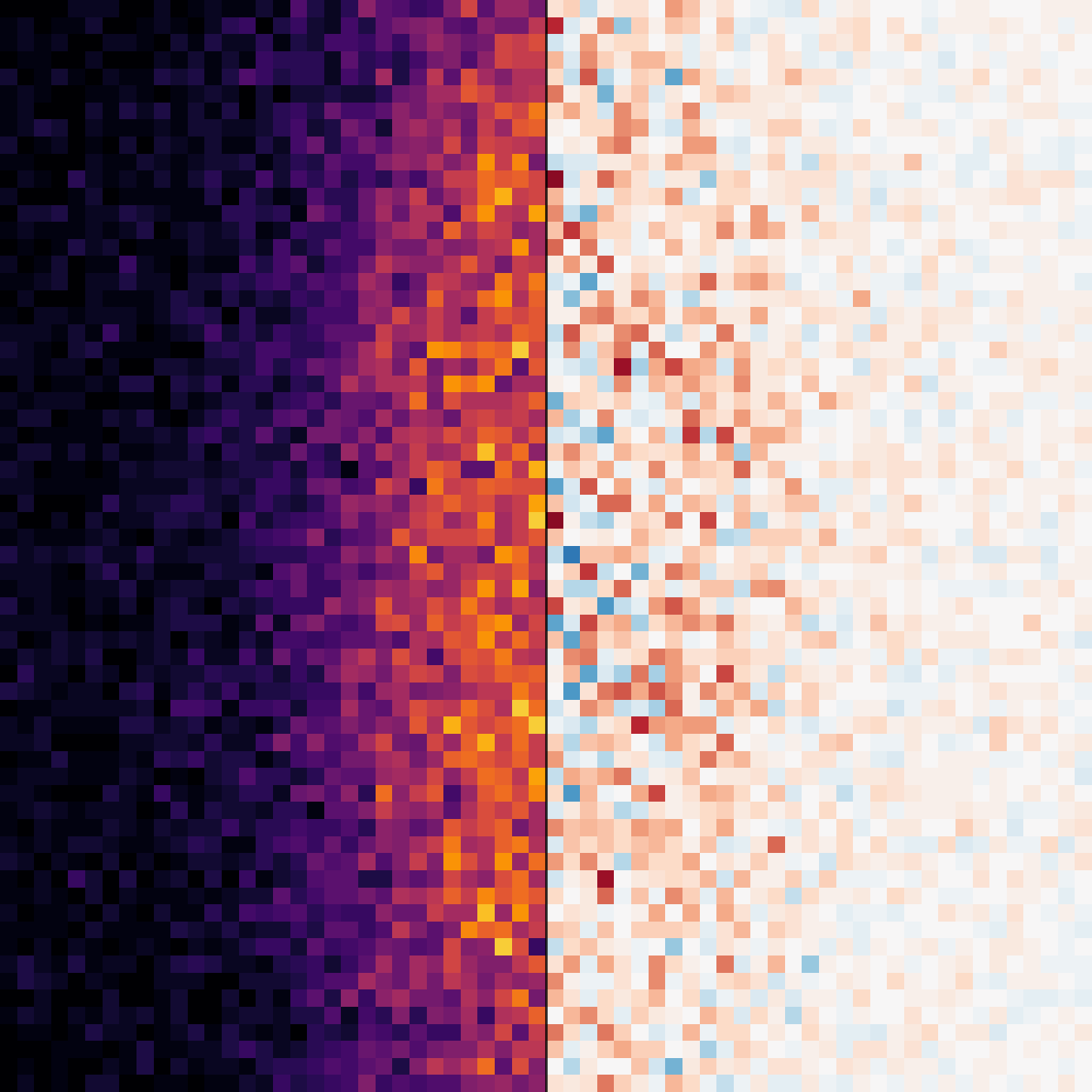}};

\node[font=\tiny, fill=white, fill opacity=0.75, text opacity=1, inner sep=\tinygap, anchor=north]
    at ($(rowB2.north west)!0.25!(rowB2.north east)$) {current};
\node[font=\tiny, fill=white, fill opacity=0.75, text opacity=1, inner sep=\tinygap, anchor=north]
    at ($(rowB2.north west)!0.75!(rowB2.north east)$) {$\Delta$ vs.\ init};

\node[tag] at ($(rowA1.south)!0.5!(rowB1.north)$) {(b)};
\node[tag] at ($(rowA2.south)!0.5!(rowB2.north)$) {(c)};
\node[tag] at ($(rowA3.south)!0.5!(rowB3.north)$) {(d)};
\node[tag] at ($(rowA4.south)!0.5!(rowB4.north)$) {(e)};
\node[tag] at ([yshift=-\tagdrop]rowB1.south) {(f)};
\node[tag] at ([yshift=-\tagdrop]rowB2.south) {(g)};
\node[tag] at ([yshift=-\tagdrop]rowB3.south) {(h)};
\node[tag] at ([yshift=-\tagdrop]rowB4.south) {(i)};

\end{tikzpicture}%
}

\caption{Neutron moderator optimization. (a) OpenMC-evaluated total current
along optimization trajectories, with Traditional, GANO, and Dataset best
references. (b--e) Initial, traditional, GANO, and our moderator geometries.
The initial panel includes the point neutron source and sensor cylinders;
the traditional baseline is rendered as an analytic cylinder.
(f--i) Corresponding OpenMC-evaluated sensor-side current densities.
For optimized designs (g--i), each panel shows current density on the left
and its change relative to the initial design on the right.}
\label{fig:overview}
\label{fig:neutron-optimization}
\end{figure}


Figure~\ref{fig:neutron-optimization}(a) shows a similar surrogate-design trend:
both shell intermediates outperform SDF and Direct. Thus, the shell interface
improves over both the non-shell SDF representation and the no-interface Direct
baseline. The full comparison with joint fine-tuning is given in
Appendix~\ref{app:finetune-comparison}.

The best optimized total current is $0.02893$, achieved with HeatSDF. This
exceeds Dataset best ($0.02542$) by $13.8\%$, GANO best ($0.02528$) by
$14.4\%$, and the traditional best ($0.02309$) by $25.3\%$.
Panels (c--e) show the optimized moderator shapes; panels (g--i) show their
OpenMC-evaluated sensor-side current densities and changes relative to the
initial design in (b,f). The current-density change plots show that our method
achieves larger improvements than either baseline.

\section{Limitations}
\label{sec:limitations}

\paragraph{Feasibility and topology.}
Fixed mesh vertices provide exact control during surface deformation, but
shell-space extraction does not automatically retain these interfaces.
Jointly changing topology and preserving hard surface constraints remains
unresolved. Volume, thickness, and containment penalties likewise do not
guarantee feasibility, and collision penalties cannot rule out all invalid
meshes. Our topology-changing evidence comes from geometric matching; the
physics results do not establish general constrained topology optimization.
Lower bracket displacement alone also does not certify stress, fatigue, or
manufacturing requirements.

\paragraph{Surrogate reliability.}
Removing a latent shape prior does not remove the training distribution's
effect on the learned forward models. Low field-prediction error need not
imply accurate shape derivatives, and optimization can exploit surrogate
errors. The different fine-tuning outcomes on DeepJEB and neutron transport
motivate checkpoint selection using optimization-relevant validation shapes
and derivative diagnostics. Independent physical evaluation remains
necessary. Where the governing problem admits symmetries, enforcing the
corresponding equivariance may improve gradients; arbitrary rotational
invariance is inappropriate when loads, sources, or boundary conditions break
that symmetry.

\paragraph{Resolution and evaluation scope.}
Surface sampling, shell-grid resolution, and diffusion or truncation scale
limit recoverable detail. Remeshing adds mesh degrees of freedom but cannot
recover information absent from the learned field, and preconditioning can
smooth away fine features. The reported tasks and initializations do not
establish robustness across arbitrary geometries or physical regimes.
Broader multi-seed studies, systematic feasibility statistics, and uncertainty
analysis are needed to characterize optimization reliability and small
differences between surrogate variants.

\section{Conclusion}
We introduced M2S-NO as the mesh-to-shell operator in a category-agnostic
inverse shape optimization framework. Its
central design separates the optimization variable from the physics-surrogate
input: explicit mesh vertices provide geometric control, while a learned
shell field provides a differentiable interface to physical-field prediction.
Preconditioned updates and remeshing support free-form deformation, and
shell-space proposals permit topology changes when the task allows them.
Geometric matching and solver-evaluated elasticity and neutron-transport
experiments demonstrate the value of this interface, including improvements
beyond the best observed training designs. The results support shell-based
surrogates as a practical component of inverse design, while highlighting
the need for constraint-preserving topology updates and forward models whose
gradients remain reliable along optimization trajectories.

\clearpage
\section*{Acknowledgments}
Anima Anandkumar is supported in part by Bren endowed chair, ONR (MURI grant
N00014-23-1-2654), and the AI2050 senior fellow program at Schmidt Sciences.

Xinyi Li and Xi Deng acknowledge support from the Pritzker Foundation through
\emph{Initiative to Advance the Integration of AI into Scientific Research and
Development} (Subaward \#ORS000086). Xi Deng also acknowledges support from
the Eric and Wendy Schmidt Fund for Strategic Innovation through
\emph{AI Accelerating Science: Neural Operators for Learning Multi-Scale
Multi-Physics Processes} (Schmidt AI2050, Award \#G-23-65858), and from the
U.S. Department of Energy SciDAC program through \emph{High-fidelity Digital
Models for Fusion Pilot Plant Design} (Award \#DE-AC02-09CH11466;
Subaward \#P-240000887). Yubo Cao and Hudson Liu acknowledge support from
Caltech's Summer Undergraduate Research Fellowships (SURF) program.

We thank Nicole Feng for sharing her perspective on heat-based signed distance
representations (HeatSDF).

\bibliography{references}

\begin{thebibliography}{24}
\providecommand{\natexlab}[1]{#1}
\providecommand{\url}[1]{\texttt{#1}}
\expandafter\ifx\csname urlstyle\endcsname\relax
  \providecommand{\doi}[1]{doi: #1}\else
  \providecommand{\doi}{doi: \begingroup \urlstyle{rm}\Url}\fi

\bibitem[Allen et~al.(2022)Allen, Lopez-Guevara, Stachenfeld, Sanchez-Gonzalez,
  Battaglia, Hamrick, and Pfaff]{allen2022physical}
Kelsey~R. Allen, Tatiana Lopez-Guevara, Kimberly Stachenfeld, Alvaro
  Sanchez-Gonzalez, Peter Battaglia, Jessica Hamrick, and Tobias Pfaff.
\newblock Physical design using differentiable learned simulators.
\newblock \emph{arXiv preprint arXiv:2202.00728}, 2022.
\newblock URL \url{https://arxiv.org/abs/2202.00728}.

\bibitem[Baque et~al.(2018)Baque, Remelli, Fleuret, and Fua]{baque2018geodesic}
Pierre Baque, Edoardo Remelli, Francois Fleuret, and Pascal Fua.
\newblock Geodesic convolutional shape optimization.
\newblock In \emph{Proceedings of the 35th International Conference on Machine
  Learning}, volume~80 of \emph{Proceedings of Machine Learning Research}, pp.\
   472--481, 2018.
\newblock URL \url{https://proceedings.mlr.press/v80/baque18a.html}.

\bibitem[Cao et~al.(2026)Cao, Deng, Xia, Gopakumar, Gray, and
  Anandkumar]{cao2026ptno}
Yubo Cao, Xi~Deng, Mengqi Xia, Vignesh Gopakumar, Ander Gray, and Anima
  Anandkumar.
\newblock {PTNO}: Training neural operators with noisy monte carlo estimates
  for particle transport problems.
\newblock \emph{arXiv preprint arXiv:2609.40090}, 2026.
\newblock \doi{10.48550/arXiv.2609.40090}.
\newblock URL \url{https://arxiv.org/abs/2609.40090}.

\bibitem[Chen et~al.(2026)Chen, Lin, Chen, Cen, Zhang, Huang, Li, and
  Sun]{chen2026aerodynamic}
Huaguan Chen, Ning Lin, Luxi Chen, Jiacheng Cen, Rui Zhang, Wenbing Huang,
  Chongxuan Li, and Hao Sun.
\newblock Optimization and generation in aerodynamics inverse design.
\newblock \emph{arXiv preprint arXiv:2602.03582}, 2026.
\newblock URL \url{https://arxiv.org/abs/2602.03582}.

\bibitem[Deng et~al.(2026)Deng, Tang, Czekanski, Bindel, and
  Marschner]{deng2026differentiable}
Xi~Deng, Maosen Tang, Michael Czekanski, David Bindel, and Steve Marschner.
\newblock Differentiable neutron transport.
\newblock \emph{ACM Transactions on Graphics}, 45\penalty0 (4):\penalty0 1--14,
  2026.
\newblock \doi{10.1145/3811296}.

\bibitem[Hong et~al.(2025)Hong, Kwon, Shin, Park, and Kang]{hong2025deepjeb}
Seongjun Hong, Yongmin Kwon, Dongju Shin, Jangseop Park, and Namwoo Kang.
\newblock Deepjeb: 3d deep learning-based synthetic jet engine bracket dataset.
\newblock \emph{Journal of Mechanical Design}, 147\penalty0 (4):\penalty0
  041703, 2025.
\newblock \doi{10.1115/1.4067089}.

\bibitem[Leibovici et~al.(2026)Leibovici, Kovachki, Ahn, Ohana, Shokar,
  Ghasemi, Akkurt, Ranade, Ashton, Kautz, and Kossaifi]{leibovici2026canto}
Daniel Leibovici, Nikola~Borislavov Kovachki, Dawon Ahn, Ruben Ohana, Ira J.~S.
  Shokar, Abouzar Ghasemi, Semih Akkurt, Rishikesh Ranade, Neil Ashton, Jan
  Kautz, and Jean Kossaifi.
\newblock {CANTO}: {CAD}-native transformer operators for {AI}-aided
  engineering.
\newblock \emph{arXiv preprint arXiv:2609.36806}, 2026.
\newblock \doi{10.48550/arXiv.2609.36806}.
\newblock URL \url{https://arxiv.org/abs/2609.36806}.

\bibitem[Li et~al.(2025)Li, Li, Kovachki, and Anandkumar]{li2025otno}
Xinyi Li, Zongyi Li, Nikola Kovachki, and Anima Anandkumar.
\newblock Geometric operator learning with optimal transport.
\newblock \emph{arXiv preprint arXiv:2507.20065}, 2025.
\newblock URL \url{https://arxiv.org/abs/2507.20065}.

\bibitem[Li et~al.(2020)Li, Kovachki, Azizzadenesheli, Liu, Bhattacharya,
  Stuart, and Anandkumar]{li2020gno}
Zongyi Li, Nikola Kovachki, Kamyar Azizzadenesheli, Burigede Liu, Kaushik
  Bhattacharya, Andrew Stuart, and Anima Anandkumar.
\newblock Neural operator: Graph kernel network for partial differential
  equations.
\newblock \emph{arXiv preprint arXiv:2003.03485}, 2020.
\newblock URL \url{https://arxiv.org/abs/2003.03485}.

\bibitem[Li et~al.(2021)Li, Kovachki, Azizzadenesheli, Liu, Bhattacharya,
  Stuart, and Anandkumar]{li2021fno}
Zongyi Li, Nikola Kovachki, Kamyar Azizzadenesheli, Burigede Liu, Kaushik
  Bhattacharya, Andrew Stuart, and Anima Anandkumar.
\newblock Fourier neural operator for parametric partial differential
  equations.
\newblock In \emph{International Conference on Learning Representations}, 2021.
\newblock URL \url{https://openreview.net/forum?id=c8P9NQVtmnO}.

\bibitem[Li et~al.(2023{\natexlab{a}})Li, Huang, Liu, and
  Anandkumar]{li2023geofno}
Zongyi Li, Daniel~Zhengyu Huang, Burigede Liu, and Anima Anandkumar.
\newblock Fourier neural operator with learned deformations for pdes on general
  geometries.
\newblock \emph{Journal of Machine Learning Research}, 24\penalty0
  (388):\penalty0 1--26, 2023{\natexlab{a}}.
\newblock URL \url{https://www.jmlr.org/papers/v24/23-0064.html}.

\bibitem[Li et~al.(2023{\natexlab{b}})Li, Kovachki, Choy, Li, Kossaifi, Otta,
  Nabian, Stadler, Hundt, Azizzadenesheli, and Anandkumar]{li2023gino}
Zongyi Li, Nikola~Borislavov Kovachki, Chris Choy, Boyi Li, Jean Kossaifi,
  Shourya~Prakash Otta, Mohammad~Amin Nabian, Maximilian Stadler, Christian
  Hundt, Kamyar Azizzadenesheli, and Anima Anandkumar.
\newblock Geometry-informed neural operator for large-scale 3d pdes.
\newblock In \emph{Advances in Neural Information Processing Systems},
  volume~36, pp.\  35836--35854, 2023{\natexlab{b}}.
\newblock \doi{10.52202/075280-1556}.

\bibitem[Lin et~al.(2025)Lin, Berner, Duruisseaux, Pitt, Leibovici, Kossaifi,
  Azizzadenesheli, and Anandkumar]{lin2025mgno}
Ryan~Y. Lin, Julius Berner, Valentin Duruisseaux, David Pitt, Daniel Leibovici,
  Jean Kossaifi, Kamyar Azizzadenesheli, and Anima Anandkumar.
\newblock Enabling automatic differentiation with mollified graph neural
  operators.
\newblock \emph{arXiv preprint arXiv:2504.08277}, 2025.
\newblock URL \url{https://arxiv.org/abs/2504.08277}.

\bibitem[Liu-Schiaffini et~al.(2024)Liu-Schiaffini, Berner, Bonev, Kurth,
  Azizzadenesheli, and Anandkumar]{li2024localno}
Miguel Liu-Schiaffini, Julius Berner, Boris Bonev, Thorsten Kurth, Kamyar
  Azizzadenesheli, and Anima Anandkumar.
\newblock Neural operators with localized integral and differential kernels.
\newblock \emph{arXiv preprint arXiv:2402.16845}, 2024.
\newblock URL \url{https://arxiv.org/abs/2402.16845}.
\newblock Accepted at 2024 International Conference on Machine Learning.

\bibitem[Nicolet et~al.(2021)Nicolet, Jacobson, and Jakob]{nicolet2021large}
Baptiste Nicolet, Alec Jacobson, and Wenzel Jakob.
\newblock Large steps in inverse rendering of geometry.
\newblock \emph{ACM Transactions on Graphics}, 40\penalty0 (6):\penalty0 1--13,
  2021.
\newblock \doi{10.1145/3478513.3480501}.

\bibitem[Oellerich \& Yamada(2025)Oellerich and Yamada]{oellerich2025levelset}
Jan Oellerich and Takayuki Yamada.
\newblock A level set topology optimization theory based on hamilton's
  principle.
\newblock \emph{International Journal for Numerical Methods in Engineering},
  126\penalty0 (17):\penalty0 e70120, 2025.
\newblock \doi{10.1002/nme.70120}.

\bibitem[Park et~al.(2019)Park, Florence, Straub, Newcombe, and
  Lovegrove]{park2019deepsdf}
Jeong~Joon Park, Peter Florence, Julian Straub, Richard Newcombe, and Steven
  Lovegrove.
\newblock Deepsdf: Learning continuous signed distance functions for shape
  representation.
\newblock In \emph{Proceedings of the IEEE/CVF Conference on Computer Vision
  and Pattern Recognition}, pp.\  165--174, 2019.

\bibitem[Rehmann et~al.(2025)Rehmann, Black, Bjorgaard, Angioi, Paleyes, Heim,
  H{\"a}fner, and Lavin]{rehmann2025surrogate}
Andrin Rehmann, Nolan Black, Josiah Bjorgaard, Alessandro Angioi, Andrei
  Paleyes, Niklas Heim, Dion H{\"a}fner, and Alexander Lavin.
\newblock Surrogate-based differentiable pipeline for shape optimization.
\newblock \emph{arXiv preprint arXiv:2511.10761}, 2025.
\newblock URL \url{https://arxiv.org/abs/2511.10761}.

\bibitem[Remelli et~al.(2020)Remelli, Lukoianov, Richter, Guillard,
  Bagautdinov, Baque, and Fua]{remelli2020meshsdf}
Edoardo Remelli, Artem Lukoianov, Stephan~R. Richter, Benoit Guillard, Timur
  Bagautdinov, Pierre Baque, and Pascal Fua.
\newblock Meshsdf: Differentiable iso-surface extraction.
\newblock In \emph{Advances in Neural Information Processing Systems},
  volume~33, pp.\  22468--22478, 2020.

\bibitem[Romano et~al.(2015)Romano, Horelik, Herman, Nelson, Forget, and
  Smith]{romano2015openmc}
Paul~K. Romano, Nicholas~E. Horelik, Bryan~R. Herman, Adam~G. Nelson, Benoit
  Forget, and Kord Smith.
\newblock {OpenMC}: A state-of-the-art monte carlo code for research and
  development.
\newblock \emph{Annals of Nuclear Energy}, 82:\penalty0 90--97, 2015.
\newblock \doi{10.1016/j.anucene.2014.07.048}.

\bibitem[Sun et~al.(2026)Sun, Miao, Huang, Chen, Wan, Zhang, and
  Sun]{sun2026gano}
Guoze Sun, Tianya Miao, Haoyang Huang, Huaguan Chen, Han Wan, Rui Zhang, and
  Hao Sun.
\newblock Geometry-aware neural optimizer for shape optimization and inversion.
\newblock \emph{arXiv preprint arXiv:2605.04474}, 2026.
\newblock URL \url{https://arxiv.org/abs/2605.04474}.
\newblock To appear in ICML 2026.

\bibitem[Wang et~al.(2024)Wang, Gocht, Ball, Mackie, Panontin, Tinguely, Raj,
  Holmes, Saltos, Johnson, and Grieve]{wang2024neutronics}
X.~Wang, R.~Gocht, J.~Ball, S.~Mackie, E.~Panontin, R.~A. Tinguely, P.~Raj,
  I.~Holmes, A.~A. Saltos, A.~Johnson, and A.~Grieve.
\newblock Neutronics simulations for the design of neutron flux monitors in
  {SPARC}.
\newblock \emph{Review of Scientific Instruments}, 95\penalty0 (8):\penalty0
  083560, 2024.
\newblock \doi{10.1063/5.0219508}.

\bibitem[Wei et~al.(2024)Wei, Dufour, Pelletier, Fua, and
  Bauerheim]{wei2024diffairfoil}
Zhen Wei, Edouard~R. Dufour, Colin Pelletier, Pascal Fua, and Micha{"e}l
  Bauerheim.
\newblock Diffairfoil: An efficient novel airfoil sampler based on latent space
  diffusion model for aerodynamic shape optimization.
\newblock In \emph{AIAA AVIATION FORUM AND ASCEND 2024}, Las Vegas, Nevada,
  2024. American Institute of Aeronautics and Astronautics.
\newblock \doi{10.2514/6.2024-3755}.

\bibitem[You et~al.(2025)You, Zhao, Zhang, Xu, and Fua]{you2025physgen}
Yingxuan You, Chen Zhao, Hantao Zhang, Ming Xu, and Pascal Fua.
\newblock Physgen: Physically grounded 3d shape generation for industrial
  design.
\newblock \emph{arXiv preprint arXiv:2512.00422}, 2025.
\newblock URL \url{https://arxiv.org/abs/2512.00422}.
\newblock Accepted to CVPR 2026.

\end{thebibliography}
\bibliographystyle{iclr2027_conference}

\clearpage
\appendix

\section{Implementation and Optimization Details}
\label{app:implementation}

\subsection{Offline target HeatSDF construction}
\label{app:pde-details}

This subsection gives the optional offline procedure used to construct target
HeatSDF fields for training and evaluation.  It is not part of the inference
path in Algorithm~\ref{alg:inverse-shape}; during inverse optimization,
$\cH_\theta$ predicts the HeatSDF directly from $(P,N)$.
The discrete Laplacian $\Delta_h$ is the second-order Cartesian Laplacian with
homogeneous Neumann boundary conditions.  A three-dimensional orthonormal
DCT-II diagonalizes it.  Along an axis of length $N_d$, its eigenvalues are
\begin{equation}
 \lambda_{d,r}=\frac{2}{h^2}
 \left[\cos\!\left(\frac{\pi r}{N_d}\right)-1\right],
 \qquad r=0,\ldots,N_d-1.
 \label{eq:appendix-neumann-spectrum}
\end{equation}
The eigenvalue for frequency $(r_1,r_2,r_3)$ is
$\lambda_{1,r_1}+\lambda_{2,r_2}+\lambda_{3,r_3}$.  Thus each backward-Euler
step is evaluated componentwise in the DCT basis as
\begin{equation}
 \widehat X^{(q+1)}_{r_1r_2r_3c}
 =\frac{\widehat X^{(q)}_{r_1r_2r_3c}}
 {1-\tau(\lambda_{1,r_1}+\lambda_{2,r_2}+\lambda_{3,r_3})}.
 \label{eq:appendix-diffusion-dct}
\end{equation}
The default values are $K=1$ and $\tau=0.05$.  When $\tau$ is selected
automatically, the code uses
\begin{equation}
 \tau=\frac{1}{2K}\left(\frac{Nh}{2}\right)^2.
 \label{eq:appendix-auto-dt}
\end{equation}

The divergence uses centered differences in the interior.  For example, its
$x$ contribution is
\begin{align}
 (D_xY)_i&=\frac{Y^x_{i+e_x}-Y^x_{i-e_x}}{2h},
 &&0<i_x<N_x-1,\\
 (D_xY)_i&=\frac{Y^x_{i+e_x}-Y^x_i}{h},
 &&i_x=0,\\
 (D_xY)_i&=\frac{Y^x_i-Y^x_{i-e_x}}{h},
 &&i_x=N_x-1.
 \label{eq:appendix-divergence}
\end{align}
with analogous terms in $y$ and $z$.  We set
$\nabla_h\cdot Y=D_xY+D_yY+D_zY$.

The Neumann Poisson equation has an additive null space.  In the DCT basis we
set the zero-frequency coefficient of $\phi$ to zero and divide every other
coefficient by the corresponding Laplacian eigenvalue.  With
$\boldsymbol r=(r_1,r_2,r_3)$,
$\Lambda_{\boldsymbol r}=\sum_{d=1}^{3}\lambda_{d,r_d}$, and
$b=\nabla_h\cdot Y$,
\begin{equation}
 \widehat\phi_{\boldsymbol r}=
 \begin{cases}
  0, & \boldsymbol r=\boldsymbol 0,\\
  \widehat b_{\boldsymbol r}/\Lambda_{\boldsymbol r}, & \text{otherwise}.
 \end{cases}
 \label{eq:appendix-poisson-dct}
\end{equation}
After cropping, the mesh-conversion pipeline places the sampled surface at zero
on average:
\begin{equation}
 \phi_i\leftarrow\phi_i-\frac{1}{S}\sum_{k=1}^{S}
 \mathcal I_h[\phi](\widetilde p_k).
 \label{eq:appendix-surface-centering}
\end{equation}
For training-aligned neural-operator targets and predictions, the alternative
centering convention subtracts the mean of $\phi$ over occupied surface
voxels, defined by $\|X_{0,i}\|_2>10^{-3}$.

\subsection{Shell-space topology proposals}
\label{app:topology-update}

Algorithm~\ref{alg:topology-update} expands the optional topology block of
Algorithm~\ref{alg:inverse-shape}. The shell-space loss omits the mesh
collision penalty. We therefore reconstruct a candidate mesh and evaluate
its full mesh-domain total loss through the composite surrogate before
acceptance. A decrease in this penalized loss is not a certificate of hard
geometric feasibility. In particular, level-set extraction does not retain
the vertex identities used to fix interfaces during mesh updates.

\begin{algorithm}[ht]
\caption{Shell-space topology update block}
\label{alg:topology-update}
\begin{algorithmic}[1]
\Require Current mesh $(V_t,F_t)$; step size $\eta_t^\phi$; geometry weight
$\lambda_{\mathrm{geo}}$; objective $\mathcal J$; frozen operators
$\cH_\theta$, $\cG_\theta$; shell-domain geometry loss
$\mathcal L_{\mathrm{geo}}^{\Phi}$
\State $(P_t,N_t)\leftarrow\operatorname{OrientedSurfaceSample}(V_t,F_t)$
\State $\widehat\phi\leftarrow\cH_\theta(P_t,N_t)$
\State $\mathcal L_{\mathrm{design}}^{\Phi}\leftarrow
  \mathcal J\!\left(\cG_\theta(\widehat\phi)\right)$
\State $\mathcal L_{\mathrm{total}}^{\Phi}\leftarrow
  \mathcal L_{\mathrm{design}}^{\Phi}+
  \lambda_{\mathrm{geo}}\mathcal L_{\mathrm{geo}}^{\Phi}(\widehat\phi)$
\State $\phi^+\leftarrow\widehat\phi-
  \eta_t^\phi\nabla_{\widehat\phi}\mathcal L_{\mathrm{total}}^{\Phi}$
\State $(V_\phi,F_\phi)\leftarrow\operatorname{MarchingCubes}(\phi^+)$
\If{$(V_\phi,F_\phi)$ changes topology and
 $\mathcal L_{\mathrm{total}}^{\mathcal M}(V_\phi,F_\phi)<
 \mathcal L_{\mathrm{total}}^{\mathcal M}(V_t,F_t)$}
 \State $(V_t,F_t)\leftarrow(V_\phi,F_\phi)$
\EndIf
\State \Return $(V_t,F_t)$
\end{algorithmic}
\end{algorithm}

\subsection{Surface sampling and coordinate normalization}
\label{app:sampling-normalization}

This subsection provides the sampling and preprocessing details omitted from
Algorithm~\ref{alg:inverse-shape}.  Faces are sampled with replacement with
probability proportional to area.  For face $\ell$, let
\begin{equation}
 A_\ell(V)=\frac12\left\|
 (V_{f_{\ell2}}-V_{f_{\ell1}})\times
 (V_{f_{\ell3}}-V_{f_{\ell1}})\right\|_2,
 \qquad
 \pi_\ell(V)=\frac{A_\ell(V)}{\sum_j A_j(V)}.
 \label{eq:appendix-face-distribution}
\end{equation}
Faces are then sampled with probability $\pi_\ell(V)$.
Specifically, $z_k\in\mathcal F$ selects row $f_{z_k}$ of $F$ with probability
$\pi_{z_k}(V)$.  To draw $\beta_k\sim\operatorname{Unif}(\Delta^2)$, we sample
$u_k,v_k\sim\mathcal U[0,1]$.  If $u_k+v_k>1$, we replace them by
$(1-u_k,1-v_k)$ and then set
\begin{equation}
 \beta_k=(1-u_k-v_k,u_k,v_k).
 \label{eq:appendix-barycentric}
\end{equation}
The face indices and barycentric coordinates are sampled without gradient
tracking.  Holding them fixed during a gradient step avoids differentiating
through discrete face selection while preserving derivatives of the sampled
positions and normals with respect to $V$.

Before input lifting, the sampled points are centered and uniformly scaled to
match the coordinate system used for training.  Mapping every mesh to the same
physical extent keeps the number of grid cells across a geometric feature
consistent between samples.

Concretely, let
\begin{align}
 \bar p&=\frac{1}{S}\sum_{k=1}^{S}p_k,\\
 d_j&=\max_k(p_{k,j}-\bar p_j)-\min_k(p_{k,j}-\bar p_j),\\
 s&=\frac{\max_j d_j}{2a},
 \qquad
 \widetilde p_k=\frac{p_k-\bar p}{\max(s,\epsilon_s)},
 \label{eq:appendix-coordinate-normalization}
\end{align}
where $a=0.9$ is the target half extent and $\epsilon_s>0$ prevents division
by zero.  During inverse optimization, $\bar p$ and $s$ are detached from the
computational graph, so gradients pass through $p_k$ but not through the
normalization statistics.
Normals are independently renormalized to unit length.

The vertex normals used for differentiable sampling are obtained by summing
unit incident-face normals and normalizing the result.  For offline
target HeatSDF construction, the implementation can instead use
$n_k=\widehat n_{z_k}^{\mathrm f}$, assigning the selected face normal directly
to every sample on that face.

\subsection{Differentiable mesh-domain geometry losses}
\label{app:geometry-losses}

This subsection specifies the soft geometry penalties used in
$\mathcal L_{\mathrm{geo}}^{\mathcal M}$.  Let
\begin{equation}
 [z]_{+,\tau}=\tau\log\left(1+\exp(z/\tau)\right)
 \label{eq:appendix-softplus}
\end{equation}
denote a smooth approximation to $\max(z,0)$.  We normalize each term by a
reference length or volume so that its weight is not determined by mesh
resolution or physical units.

\paragraph{Self-intersection and collision loss.}
Let $\mathcal C(V,F)$ be a broad-phase set of pairs of triangles that do not
share a vertex and whose expanded axis-aligned bounding boxes are within a
collision margin $d_{\mathrm{col}}$.  For each pair $(a,b)\in\mathcal C$, we
estimate the separation $d_{ab}$ by a soft minimum over differentiable
vertex--triangle and edge--edge distances:
\begin{equation}
 d_{ab}=-\tau_d\log\sum_{r\in\mathcal R_{ab}}
 \exp\left(-d_{ab,r}/\tau_d\right),
 \label{eq:appendix-soft-triangle-gap}
\end{equation}
where $\mathcal R_{ab}$ contains the closest-point primitives for the two
triangles.  The candidate pair list and active closest-point features are held
fixed while differentiating one mesh update and recomputed afterward.  The
collision energy is
\begin{equation}
 \mathcal L_{\mathrm{col}}^{\mathcal M}
 =\frac{1}{|\mathcal C|}\sum_{(a,b)\in\mathcal C}
 \left(\frac{[d_{\mathrm{col}}-d_{ab}]_{+,\tau_d}}
 {d_{\mathrm{col}}}\right)^2 .
 \label{eq:appendix-collision-loss}
\end{equation}
When $\mathcal C$ is empty, we set this term to zero.  Otherwise it is
negligible when the triangles are farther apart than the safety margin and
grows as they approach or intersect.  The edge--edge terms are important for
detecting crossings that place no triangle vertex inside the other triangle.
This is a differentiable barrier rather than an exact certificate of
intersection-free geometry; an exact triangle-intersection test can therefore
be applied when validating a proposed mesh or accepting a topology proposal.

\paragraph{Thickness loss.}
For a surface sample $p_i$ with outward unit normal $n_i$, we cast a ray from
$p_i$ in the inward direction $-n_i$ and let $q_i$ be the first valid
intersection with a nonadjacent triangle on the opposite surface sheet.  The
local thickness estimate is
\begin{equation}
 t_i=-n_i^{\mathsf T}(q_i-p_i).
 \label{eq:appendix-thickness-estimate}
\end{equation}
Ray-hit identities are discrete, so they are held fixed during a gradient
step; the intersection position remains differentiable with respect to the
vertices of the selected triangles.  Samples without a valid nonadjacent hit
are omitted from the thickness average.  For a lower bound $t_{\min}$ and,
optionally, an upper bound $t_{\max}$, we use
\begin{equation}
 \mathcal L_{\mathrm{thick}}^{\mathcal M}
 =\frac{1}{|\mathcal I|}\sum_{i\in\mathcal I}
 \left[
 \left(\frac{[t_{\min}-t_i]_{+,\tau_t}}{t_{\min}}\right)^2
 +\left(\frac{[t_i-t_{\max}]_{+,\tau_t}}{t_{\max}}\right)^2
 \right],
 \label{eq:appendix-thickness-loss}
\end{equation}
where the second term is omitted when no upper bound is prescribed.  Excluding
adjacent triangles prevents the local one-ring from being mistaken for the
opposite surface sheet.

\paragraph{Volume loss.}
Assume that $F$ is consistently oriented.  The signed volume enclosed by the
closed mesh is computed directly from the triangular faces:
\begin{equation}
 V_{\mathcal M}(V,F)=\frac{1}{6}\sum_{(i,j,k)\in F}
 v_i^{\mathsf T}(v_j\times v_k).
 \label{eq:appendix-mesh-volume}
\end{equation}
For a prescribed interval $[V_{\min},V_{\max}]$, the normalized penalty is
\begin{equation}
 \mathcal L_{\mathrm{vol}}^{\mathcal M}
 =\left(\frac{[V_{\min}-V_{\mathcal M}]_{+,\tau_V}}
 {V_{\mathrm{ref}}}\right)^2
 +\left(\frac{[V_{\mathcal M}-V_{\max}]_{+,\tau_V}}
 {V_{\mathrm{ref}}}\right)^2,
 \label{eq:appendix-volume-loss}
\end{equation}
where $V_{\mathrm{ref}}$ is the initial or target volume.  For an equality
constraint $V_{\mathcal M}=V^*$, we replace~\eqref{eq:appendix-volume-loss}
with $((V_{\mathcal M}-V^*)/V_{\mathrm{ref}})^2$.  Equation~\eqref{eq:appendix-mesh-volume}
is differentiable with respect to vertex positions for fixed connectivity and
is evaluated again after remeshing or topology extraction.

The mesh-domain geometry loss combines these terms with the package-box and
any fixed-interface terms as
\begin{equation}
 \mathcal L_{\mathrm{geo}}^{\mathcal M}
 =w_{\mathrm{col}}\mathcal L_{\mathrm{col}}^{\mathcal M}
 +w_{\mathrm{thick}}\mathcal L_{\mathrm{thick}}^{\mathcal M}
 +w_{\mathrm{vol}}\mathcal L_{\mathrm{vol}}^{\mathcal M}
 +w_{\mathrm{box}}\mathcal L_{\mathrm{box}}^{\mathcal M}.
 \label{eq:appendix-mesh-geometry-loss}
\end{equation}
The collision term is used only for mesh-space updates, as stated in
Section~\ref{par:geometry-constraints}.  All three penalties are soft terms;
their weights and final constraint residuals should be reported with the
corresponding experiments.

\subsection{Preconditioner implementation and optimization configuration}
\label{app:inverse-details}

This subsection instantiates the abstract operators and update in
Algorithm~\ref{alg:inverse-shape}.  The frozen map $\cH_\theta$ is the
mGNO--FNO that consumes oriented surface samples $(P,N)$ and predicts the
HeatSDF field $\phi$.  If the implementation uses a regular latent grid, the
mGNO applies its learned mollified kernel directly to the oriented samples and
latent query locations; no fixed splatting-to-grid preprocessing step is used.
This learned lifting is distinct from the fixed grid splatting used only to
construct offline HeatSDF targets in Section~\ref{app:pde-details}.  All model
parameters are frozen, but gradients are retained with respect to mesh-derived
inputs.

In addition to the target discrepancy, the implementation optionally penalizes
nonzero predicted SDF values on target-surface samples $Q^*$:
\begin{equation}
 \mathcal L(V)=w_Td_{\mathcal Y}(\widehat y,y^*)
 +\frac{w_S}{|Q^*|}\sum_{q\in Q^*}
 \rho\!\left(\mathcal I_h[\widehat\phi](q)\right),
 \label{eq:appendix-surface-loss}
\end{equation}
where $\rho$ is squared, absolute, or smooth-$\ell_1$ error.  The default is
$w_S=0$, leaving only the target-volume discrepancy.

\paragraph{Differential-coordinate realization.}
We implement the smoothing preconditioner in~\eqref{eq:mesh-preconditioner}
using the differential-coordinate parameterization introduced by Large Steps
in Inverse Rendering of Geometry~\cite{nicolet2021large}.  Rather than forming
the dense matrix $P_F$, we encode the vertices as
\begin{equation}
 U=M_FV
 \qquad\text{and decode them by}\qquad
 V=M_F^{-1}U.
 \label{eq:appendix-differential-coordinates}
\end{equation}
For fixed $F$, differentiation through the linear solve gives
\begin{equation}
 \nabla_U\mathcal L=M_F^{-T}\nabla_V\mathcal L.
 \label{eq:appendix-u-gradient}
\end{equation}
Consequently, a gradient step in $U$ produces
\begin{align}
 U^+&=U-\eta M_F^{-T}\nabla_V\mathcal L,\\
 V^+&=M_F^{-1}U^+
   =V-\eta M_F^{-1}M_F^{-T}\nabla_V\mathcal L,
 \label{eq:appendix-preconditioner-equivalence}
\end{align}
which is exactly the preconditioned vertex update in
Equation~\eqref{eq:preconditioned-update}.  The two sparse linear solves apply
the smoothing preconditioner without explicitly constructing its dense
inverse.  In this sense, Large Steps is the implementation used for our
Laplacian smoothing preconditioner, rather than the name of the optimization
objective or overall method.

The implementation also supports the alpha-parameterized factor
\begin{equation}
 M_F=(1-\alpha)I_n+\alpha L_F,
 \qquad \alpha\in[0,1),
 \label{eq:appendix-alpha-matrix}
\end{equation}
with default $\alpha=0.95$.  Up to the scalar factor $1-\alpha$, this is
equivalent to $I_n+\lambda_PL_F$ with
$\lambda_P=\alpha/(1-\alpha)$; the resulting global scale can be absorbed into
the learning rate.  Decoding is performed by conjugate gradients by default,
with Cholesky factorization as an option.

The code applies AdamUniform to $U$, yielding an adaptive extension of the
plain preconditioned-gradient step derived above.  It uses initial learning
rate $5\times10^{-4}$, multiplied by $0.8$ every 15 iterations, and clips the
$U$-gradient to unit Euclidean norm.  The surface variables $\xi$ are redrawn
every five iterations.  After each optimizer update, the decoded vertices are
translated to restore the current reference centroid, then re-encoded as
$U=M_FV$.

\subsection{Botsch--Kobbelt Remesher}
\label{app:remeshing}

During shape optimization, vertex motion can stretch some triangles and
compress others.  The resulting mesh may contain very long edges, clusters of
short edges, irregular vertex valences, and poorly shaped triangles.  These
defects reduce geometric resolution in stretched regions, waste vertices in
compressed regions, and can make both surface sampling and the mesh Laplacian
less reliable.  Remeshing aims to replace this degraded discretization by an
approximately uniform, well-shaped triangulation with target edge length
$h_R$, while preserving the represented surface and its topology as closely
as possible.  Thus remeshing changes the mesh discretization, not the intended
optimized geometry.

The operator $\cR_{\mathrm{BK}}$ in Algorithm~\ref{alg:inverse-shape} is the
local isotropic remesher of Botsch and Kobbelt distributed with the reference
implementation of Nicolet et al.~\cite{nicolet2021large}.  It targets three
complementary mesh properties: edge lengths concentrated near $h_R$, vertex
valences near six in regular interior regions, and triangles with improved
aspect ratios.  It assumes a closed, edge-manifold triangle mesh.  For a target
edge length $h_R$, one BK iteration applies four local stages in order:
\begin{enumerate}
 \item split edges longer than $1.4h_R$ by inserting their midpoints;
 \item collapse edges shorter than $0.7h_R$, provided the collapse preserves
 topology and orientation and does not create incident edges longer than
 $1.4h_R$;
 \item flip interior edges whenever the flip decreases the total deviation of
 the four incident vertex valences from the regular triangular-mesh valence
 six; and
 \item move non-feature vertices toward their one-ring barycenters only in the
 local tangent plane, then project them to the pre-remeshing surface.
\end{enumerate}
Splitting and collapsing redistribute geometric resolution, flipping improves
connectivity and regularity, and tangential relaxation improves element shape
without intentionally moving vertices in the normal direction.  Projection is
enabled in both that reference experiment and our implementation,
preventing the relaxation step from drifting away from the pre-remeshing
surface.  Together, these operations seek a better-conditioned triangulation
without treating remeshing itself as an additional shape update.  The four
stages are repeated $J_R=5$ times.

The reference remeshing experiment of Nicolet et al. invokes this operator at a
prescribed optimization iteration with
$h_R=\tfrac12\,\operatorname{mean}_{(i,j)\in E_F}\|V_i-V_j\|_2$.  Our inverse
optimization uses the same remesher but a different trigger and target policy,
defined below.

Let $E_F$ be the unique undirected edge set and
$\ell_{ij}=\|V_i-V_j\|_2$.  The uniform target edge length is fixed from the
initial mesh as
\begin{equation}
 h_R=\operatorname{median}\{\ell_{ij}:(i,j)\in E_F\}.
 \label{eq:appendix-remesh-target}
\end{equation}
At each enabled remeshing check, let $q_{10}$ and $q_{90}$ be the $0.1$ and
$0.9$ quantiles of the current edge lengths.  Botsch--Kobbelt remeshing is
triggered when
\begin{equation}
 q_{10}<0.5h_R
 \qquad\text{or}\qquad
 q_{90}>1.5h_R.
 \label{eq:appendix-remesh-trigger}
\end{equation}
Since remeshing changes the number and connectivity of vertices, the old
preconditioner factor $M_F$ and differential-coordinate optimizer moments are
invalid.  We therefore rebuild $L_F$ and $M_F$, set $U=M_FV$, reset
AdamUniform, and redraw $\xi$ on the new surface, as indicated in
Algorithm~\ref{alg:inverse-shape}.

\section{Experimental details and additional results}


\subsection{StableSDF fitting for the arch-augmented bracket}
\label{app:gano-stablesdf}

GANO uses StableSDF as its SDF auto-decoder.  Before latent-space optimization,
we therefore optimized the StableSDF latent code to fit the initial
arch-augmented bracket.  The fitting procedure failed to represent this
composite geometry, so it did not produce a valid StableSDF initialization for
GANO and no GANO trajectory is reported for this case in
Table~\ref{tab:deepjeb-fem-results}.  We consequently treat the
arch-augmented bracket as out of distribution with respect to the learned
StableSDF/GANO shape prior.

\subsection{Frozen and Fine-Tuned Optimization Comparison}
\label{app:finetune-comparison}

\begin{figure*}[t]
  \centering
  \includegraphics[width=0.24\textwidth]{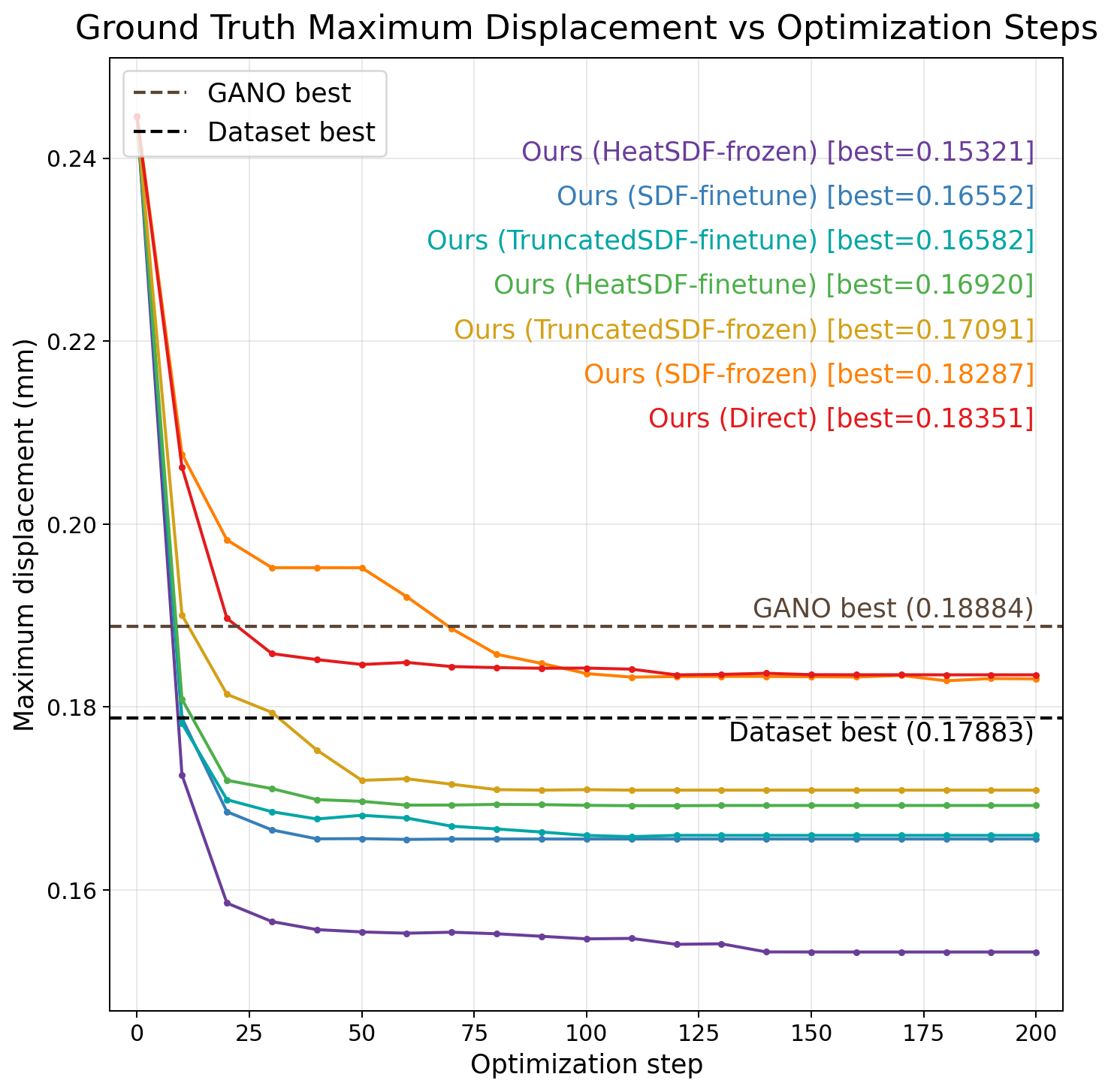}\hfill
  \includegraphics[width=0.24\textwidth]{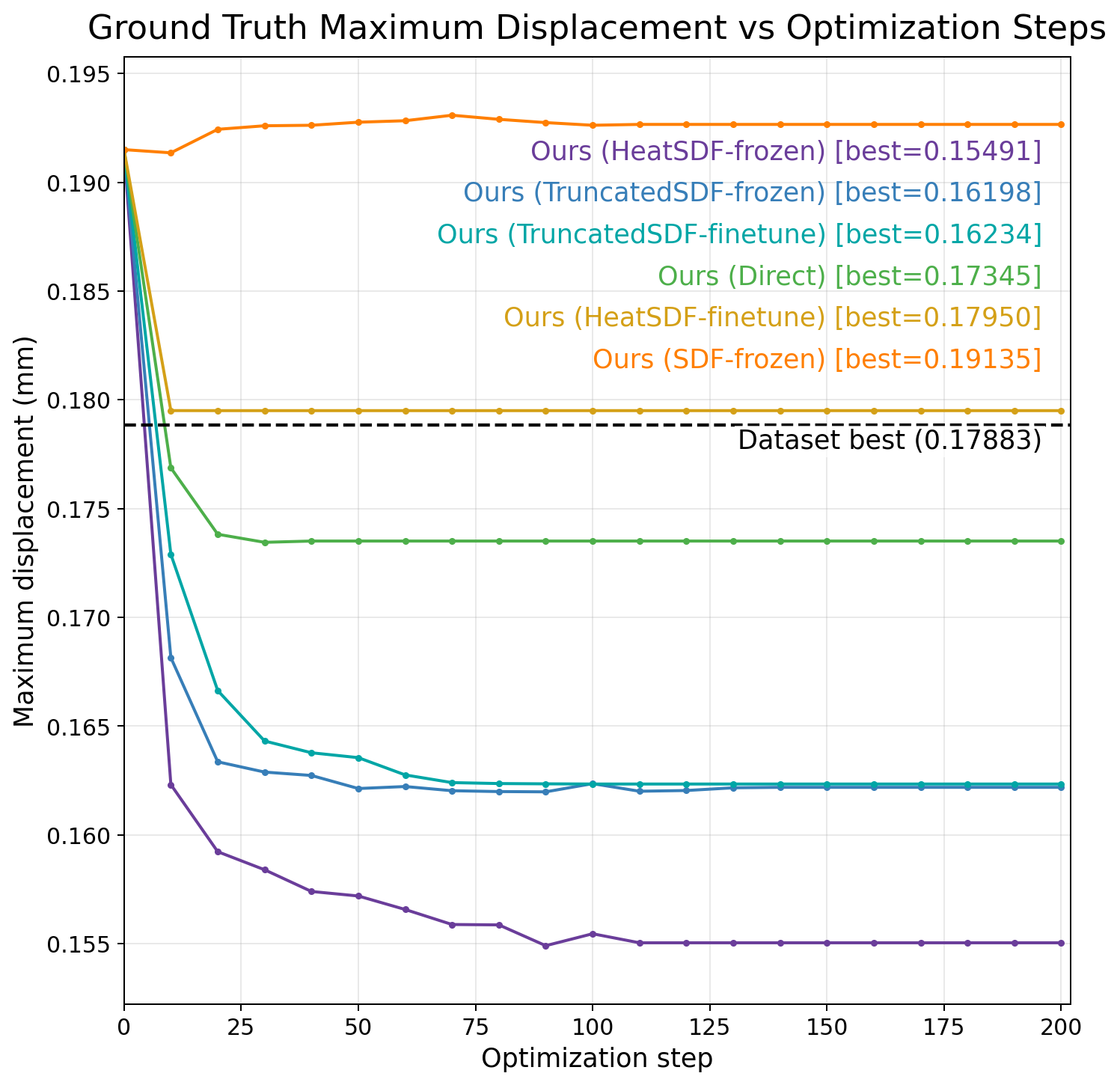}\hfill
  \includegraphics[width=0.48\textwidth]{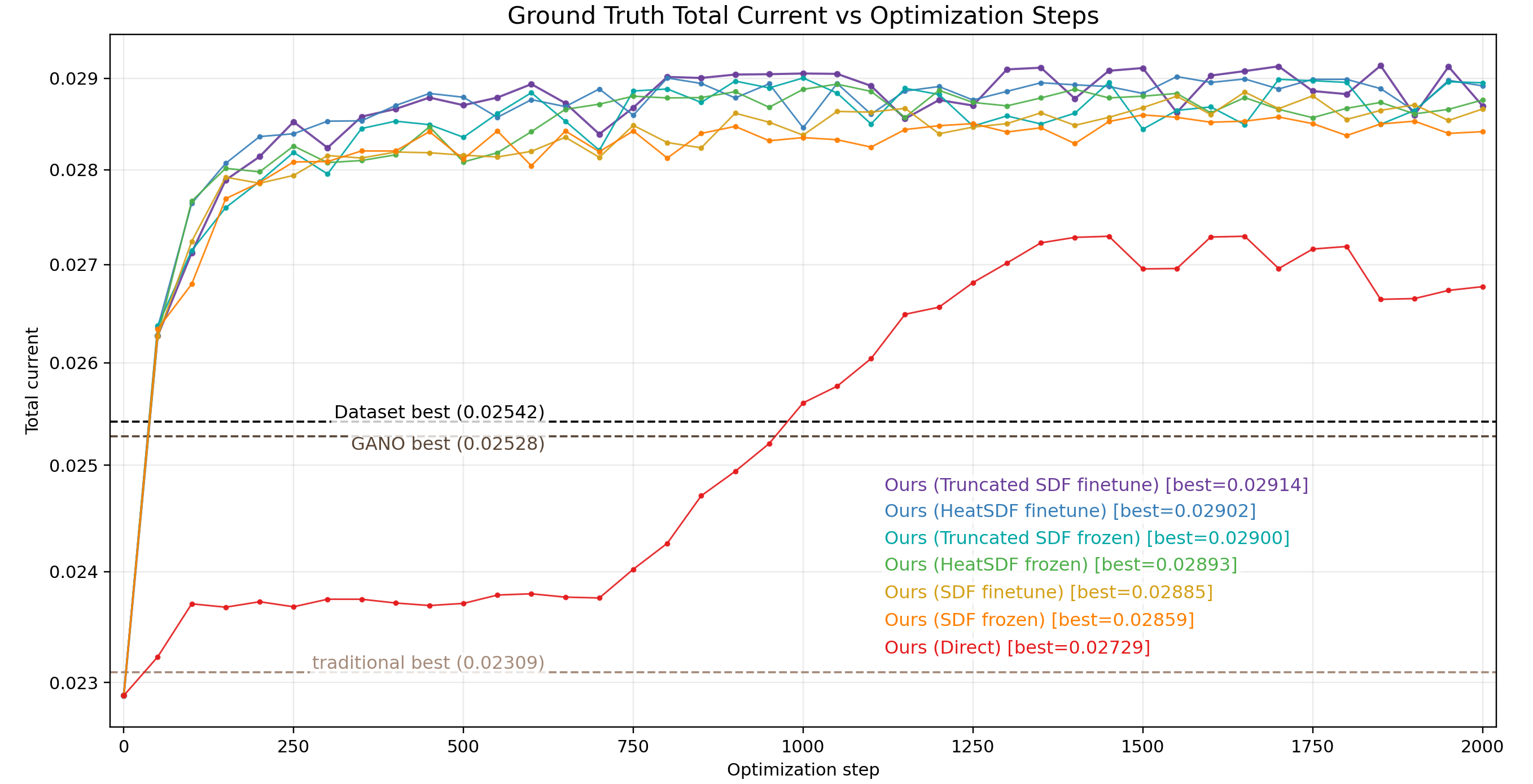}
  \caption{Ground-truth optimization trajectories for the in-distribution
  DeepJEB bracket (left), the arch-augmented OOD bracket (middle), and the
  neutron moderator (right). Each panel compares Direct, SDF, TruncatedSDF,
  and HeatSDF with fixed weights and with joint fine-tuning. Lower is better
  for the two displacement plots; higher is better for neutron current.}
  \label{fig:finetune-losses}
\end{figure*}

The main text uses independently pretrained M2S-NO and PDE-NO models with all
weights fixed during inverse optimization. For completeness,
Figure~\ref{fig:finetune-losses} compares this setting with joint fine-tuning
using the physical-field prediction loss. The curves are evaluated with the
corresponding ground-truth FEM or OpenMC objective, rather than with the
training loss. In the fixed-weight setting, both shell representations
consistently outperform the Direct and SDF variants across the three physics
tasks. Joint fine-tuning changes the ranking of some variants, but the shell
representations remain competitive; these comparisons are not used for the
main-text claims.

\section{Neutron Current Optimization Details}
\label{app:neutron-current}

We first generate $10{,}000$ base moderator geometries with CSG.  Because
topology optimization is outside the scope of this experiment, construction
uses only union operations, and we retain only connected shapes with the
prescribed topology that fully enclose the cylindrical sensor.  The moderator
material is fixed for all designs.  For each base geometry, we produce ten
configurations by rotation, translation, and scaling while preserving these
constraints, yielding $100{,}000$ moderator cases in total.

For every case, OpenMC generates a noisy transport label consisting of the
incoming middle-energy-group current density sampled on a receiver grid over
the cylindrical sensor side.  We use the noisy-label generation and forward
model training protocol from PTNO~\citep{cao2026ptno} to train the
problem-specific PDE-NO on this dataset.  The trained PDE-NO predicts the
receiver-grid current density from the volumetric moderator representation;
the optimization objective in Equation~\eqref{eq:neutron-current-objective}
integrates this prediction over the sensor side.  Final optimized candidates
are evaluated independently with OpenMC.

\end{document}